%% file: main.tex
\documentclass[sigconf]{acmart}
\AtBeginDocument{%
  }

\copyrightyear{2026}
\acmYear{2026}
\setcopyright{cc}
\setcctype{by}
\acmConference[KDD '26]{Proceedings of the 32nd ACM SIGKDD Conference on Knowledge Discovery and Data Mining V.2}{August 09--13, 2026}{Jeju Island, Republic of Korea}
\acmBooktitle{Proceedings of the 32nd ACM SIGKDD Conference on Knowledge Discovery and Data Mining V.2 (KDD '26), August 09--13, 2026, Jeju Island, Republic of Korea}
\acmDOI{10.1145/3770855.3817614}
\acmISBN{979-8-4007-2259-2/2026/08}
\usepackage{algorithm}
\usepackage{algpseudocode} 
\usepackage{amsmath}       
\usepackage{booktabs,tabularx}
\usepackage{booktabs,multirow}
\usepackage{graphicx} 
\usepackage{caption}
\usepackage{placeins}
\usepackage[table]{xcolor}
\usepackage{array}
\usepackage{makecell}
\usepackage{xcolor}  
\usepackage{enumitem}

\definecolor{percentgray}{RGB}{100,100,100}  
\newcommand{\pct}[1]{{\scriptsize\textcolor{percentgray}{(#1)}}}

\newcolumntype{Y}{>{\centering\arraybackslash}X}

\begin{document}

\title[TS-Memory: Plug-and-Play Memory for Time Series Foundation Models]{TS-Memory: Plug-and-Play Memory for \\ Time Series Foundation Models}

\author{Sisuo Lyu}
\authornote{Equal Contribution.}
\affiliation{%
  \institution{The Hong Kong University of Science and Technology (Guangzhou)}
  \city{Guangzhou}
  \country{China}
}
\email{sisuolyu@outlook.com}

\author{Siru Zhong}
\authornotemark[1]
\affiliation{%
  \institution{The Hong Kong University of Science and Technology (Guangzhou)}
  \city{Guangzhou}
  \country{China}
}
\email{siruzhong@outlook.com}

\author{Tiegang Chen}
\affiliation{%
  \institution{Tencent}
  \city{Shenzhen}
  \country{China}
}
\email{steelchen@tencent.com}

\author{Weilin Ruan}
\affiliation{%
  \institution{The Hong Kong University of Science and Technology (Guangzhou)}
  \city{Guangzhou}
  \country{China}
}
\email{rwlinno@gmail.com}

\author{Qingxiang Liu}
\affiliation{%
  \institution{The Hong Kong University of Science and Technology (Guangzhou)}
  \city{Guangzhou}
  \country{China}
}
\email{qingxiangliu737@gmail.com}

\author{Taiqiang Lv}
\affiliation{%
  \institution{Tencent}
  \city{Shenzhen}
  \country{China}
}
\email{fielixlv@tencent.com}

\author{Qingsong Wen}
\affiliation{%
  \institution{Squirrel Ai Learning}
  \city{Bellevue}
  \state{Washington}
  \country{USA}
}
\email{qingsongedu@gmail.com}

\author{Raymond Chi-Wing Wong}
\authornote{Corresponding Author.}
\affiliation{%
  \institution{The Hong Kong University of Science and Technology}
  \city{Hong Kong}
  \country{China}
}
\email{raywong@cse.ust.hk}

\author{Yuxuan Liang}
\authornotemark[2]
\affiliation{%
  \institution{The Hong Kong University of Science and Technology (Guangzhou)}
  \city{Guangzhou}
  \country{China}
}
\email{yuxliang@outlook.com}
\renewcommand{\shortauthors}{Sisuo Lyu et al.}

\begin{abstract}

\textit{Time Series Foundation Models} (TSFMs) achieve strong zero-shot forecasting through large-scale pre-training, but adapting them to downstream domains under distribution shift remains challenging. Existing solutions face a trade-off: \textit{Parametric Adaptation} can cause catastrophic forgetting and requires costly multi-domain maintenance, while \textit{Non-Parametric Retrieval} improves forecasts but incurs high inference latency due to datastore search. We propose \textit{Parametric Memory Distillation} and implement it as \textbf{TS-Memory}, a lightweight memory adapter that augments frozen TSFMs. 
TS-Memory is trained in two stages. First, we construct an offline, retrieval-leakage-safe $k$NN teacher that synthesizes confidence-aware quantile targets from retrieved futures. Second, we distill this retrieval-induced distributional correction into a lightweight memory adapter via confidence-gated supervision.
During inference, TS-Memory fuses memory and backbone predictions with constant-time overhead, enabling retrieval-free deployment. Experiments across diverse TSFMs and benchmarks demonstrate consistent improvements in both point and probabilistic forecasting over representative adaptation methods, with efficiency comparable to the frozen backbone. Code: https://github.com/sisuolv/TS-Memory.

\end{abstract}



    \maketitle
\input{contents/1_introduction}
\input{contents/2_related_work}
\input{contents/3_method}

\input{contents/4_experiment}

\input{contents/5_conclusion}

\bibliographystyle{ACM-Reference-Format}
\bibliography{main}

\clearpage
\appendix

\input{contents/Appendix}

\end{document}

%% file: contents/1_introduction.tex
\section{Introduction}

Time series forecasting is a cornerstone of decision-making in critical domains such as energy, healthcare, and supply chains~\cite{rezaei2021stock,tzelepi2023deep,sun2022accurate,idrees2019prediction,kiyasseh2021clocs,liu2025towards,lei2026hierarchical}. Recently, the field has witnessed a paradigm shift from training task-specific models from scratch~\cite{zhou2021informer,wu2021autoformer,zhou2022fedformer,wu2023timesnet,liu2023itransformer,lyu2025occamvts} to \textit{Time Series Foundation Models} (TSFMs)~\cite{das2024decoder,goswami2024moment,liu2024timer,liu2025sundial,xiao2025timefound,graf2025flowstate}, which are pre-trained on diverse corpora to learn transferable temporal representations. However, a critical gap~\cite{hinder2024one,wen2023onenet,zhao2025proactive,jin2022domain,he2023domain,zhonglearning,zhong2026learning} remains between pre-training and deployment: real-world domain shifts (e.g., scale changes, seasonal discrepancies, sensor variations) can severely degrade the performance of frozen TSFMs.

To bridge this gap, existing adaptation strategies primarily fall into two paradigms (As shown in \autoref{fig:paradigm}):




\begin{figure}[t]
  \centering
  \includegraphics[width=\linewidth]{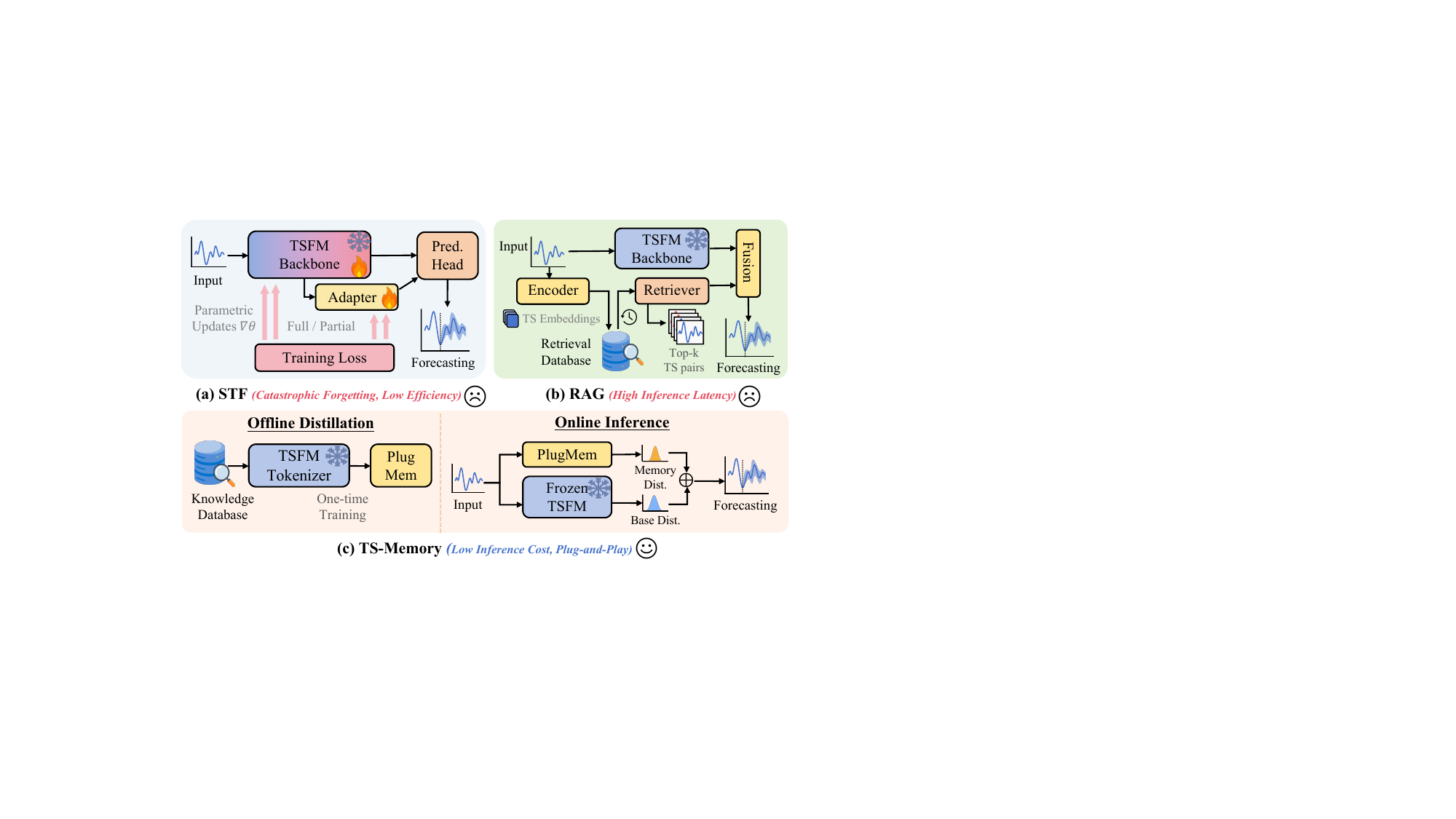}

    \caption{Comparison of TSFM adaptation paradigms: \textbf{(a)} Parametric Adaptation; \textbf{(b)} Non-Parametric Retrieval; \textbf{(c)} Parametric Memory Distillation (Ours).}
  
  \Description{Comparison of TSFM adaptation paradigms}
  \label{fig:paradigm}

\end{figure}



\begin{enumerate}[leftmargin=*]
    \item \textbf{Parametric Adaptation:} Methods like full or parameter efficient fine-tuning adapt model weights to the target domain~\cite{qiao2025multi,zhao2025less,ruan2025st}. Despite their effectiveness, this approach requires maintaining a distinct model copy for each domain, leading to exorbitant storage and maintenance costs. Moreover, continuous updates to frozen backbones risk catastrophic forgetting~\cite{fu2025financial,beichter2025decision,yu2025time} of general pre-training knowledge~\cite{ning2025mode,lee2025lightweight,ye2023mlora,karaouli2025time}.

    \item \textbf{Non-Parametric Retrieval:} To preserve the frozen backbone, methods like Metric Matching~\cite{vinyals2016matching,snell2017prototypical,sung2018learning} and Retrieval Augmented Generation (RAG) retrieve historical instances from an external database during inference~\cite{lewis2020retrieval,jing2022retrieval,liu2024retrieval,yang2025timerag,han2025retrieval,ruan2025rastretrievalaugmentedspatiotemporal}. While this avoids weight updates, it places heavy index maintenance and kNN queries on the \textit{critical inference path}. The latency scales linearly with the size of the external knowledge base, making such methods prohibitive for real-time or edge deployment~\cite{fan2024survey,ning2025ts,jin2025ragcache,zhu2024accelerating,asai2024reliable,lei2025game} under tight latency constraints.
\end{enumerate}



Existing methods face a dilemma between \textbf{maintenance} (per-domain model adaptation) and \textbf{inference} efficiency (runtime search cost). This raises a key question: \textit{Can we capture the robust adaptability of retrieval-based methods while maintaining the constant-time inference and deployment simplicity of parametric adapters?}




We propose a third paradigm: \textbf{Parametric Memory Distillation} (Table~\ref{tab:match_rag_tsmemory}). Our core insight is that the "knowledge" provided by online retrieval, namely the predictive distribution implied by similar inputs, can be compiled offline into a compact neural module. By shifting retrieval cost to training, we decouple retrieval benefits from runtime costs. Concretely, we introduce \textbf{TS-Memory}, a plug-and-play adapter for frozen TSFMs. Unlike RAG, which queries a datastore at every inference step, TS-Memory learns to approximate the teacher distribution from an offline kNN retriever. During inference, it serves as an internalized memory of domain patterns that fuses seamlessly with the TSFM's predictions.

This approach satisfies three critical deployment requirements: (1) \textbf{Zero Inference Search:} It eliminates dependencies on external vector databases during runtime; (2) \textbf{Constant Time Complexity:} Inference latency is $O(1)$, independent of the historical data size; (3) \textbf{Non-Destructive Adaptation:} It bypasses the risk of catastrophic forgetting by keeping the backbone frozen. As visualized in Figure~\ref{fig:paradigm}, TS-Memory bridges the gap between parametric adaptation and non-parametric retrieval. Rather than storing or querying retrieved instances at test time, we distill the \emph{distributional corrections} that retrieval induces on the frozen TSFM's base forecast into a lightweight neural memory module. This preserves the frozen backbone to prevent forgetting, adapts to local domains as RAG does, and operates with the speed of a standard forward pass.

Technically, TS-Memory operates in two phases. First, during an offline stage, we construct a \textit{privileged supervision} by retrieving $k$-nearest neighbors for training instances and aggregating their future trajectories into \emph{quantile targets}. This process captures the non-parametric uncertainty inherent in the data. Second, we train a lightweight parametric memory module to predict these teacher quantiles directly from the input context. To ensure robustness, we employ a \textit{confidence-gated memory distillation} that selectively transfers knowledge only when retrieval provides reliable signals. During inference, the memory module functions as a plug-and-play addition to the frozen TSFM, fused via linear interpolation to enhance adaptability without incurring any search latency.

Our contributions are summarized as follows:
\begin{itemize}[leftmargin=*]
    \item \textbf{Paradigm Shift:} We propose the Parametric Memory Distillation framework for TSFMs, transforming expensive online retrieval into a learnable, offline memory mechanism.
    
    \item \textbf{TS-Memory:} We design a lightweight module that distills non-parametric kNN distributions into parametric representations, featuring a novel distribution fusion interface for frozen TSFMs.
    
    \item \textbf{Efficiency \& Performance:} Extensive experiments demonstrate that TS-Memory outperforms representative methods from both paradigms in accuracy while significantly reducing inference latency, offering a superior trade-off for practical deployment.
\end{itemize}

\input{tables/LLMvsRAGvsMemory}

%% file: tables/LLMvsRAGvsMemory.tex


\begin{table}[t]
  \centering

  \caption{Comparison of TSFM adaptation paradigms. TS-Memory combines low latency with low storage cost.}
  \label{tab:match_rag_tsmemory}

  \footnotesize
  \setlength{\tabcolsep}{3pt}
  \renewcommand{\arraystretch}{1.2}
  \scalebox{0.86}{
  \begin{tabular}{lcccccc}
    \toprule
    Paradigm & Inference & Deployment & Latency & Backbone & Plug and\\
    & Search? & Cost & Scaling & Status &Play?\\
    \midrule
    Parametric Adaptation & No & High (Model Copies) & $O(1)$ & Updated & No\\
    Non-Parametric Retrieval & Yes & High (Vector DB) & $O(|\mathcal{D}|)$ & Frozen & Yes\\
    \rowcolor{gray!10} \textbf{Parametric Distillation} & \textbf{No} & \textbf{Low (Tiny Module)} & $\mathbf{O(1)}$ & \textbf{Frozen} & \textbf{Yes}\\
    \bottomrule
  \end{tabular}
  }

\end{table}

%% file: contents/2_related_work.tex
\section{Related Work}
\noindent\textbf{Time Series Foundation Models (TSFMs).} Research has moved from task-specific training to pre-training TSFMs on large corpora for broad transferability. Early work~\cite{rasul2023lag,garza2023timegpt} has advanced to elaborate architectures with patch-based modeling~\cite{ekambaram2024tiny,das2024decoder,goswami2024moment,liu2024timer,liu2024timerxl,liu2025sundial,xiao2025timefound,graf2025flowstate}, sparse mixture-of-experts~\cite{shi2024time,liu2024moirai}, and discretization-based probabilistic objectives~\cite{ansari2024chronos,woo2024unified,ansari2025chronos}. Concurrently, adapting general-purpose large models (LLMs/VLMs) to time series via re-programming or prompting remains active~\cite{zhou2023one,jin2023time,niu2025langtime,sun2025adapting,zhongtime,lyu2025occamvts}. While TSFMs exhibit strong generalization, purely parametric variants perform poorly under severe distribution shifts in downstream domains, demanding effective adaptation strategies~\cite{hinder2024one,jin2022domain,he2023domain}.


\begin{figure*}[t]
  \centering
  \includegraphics[width=\textwidth]{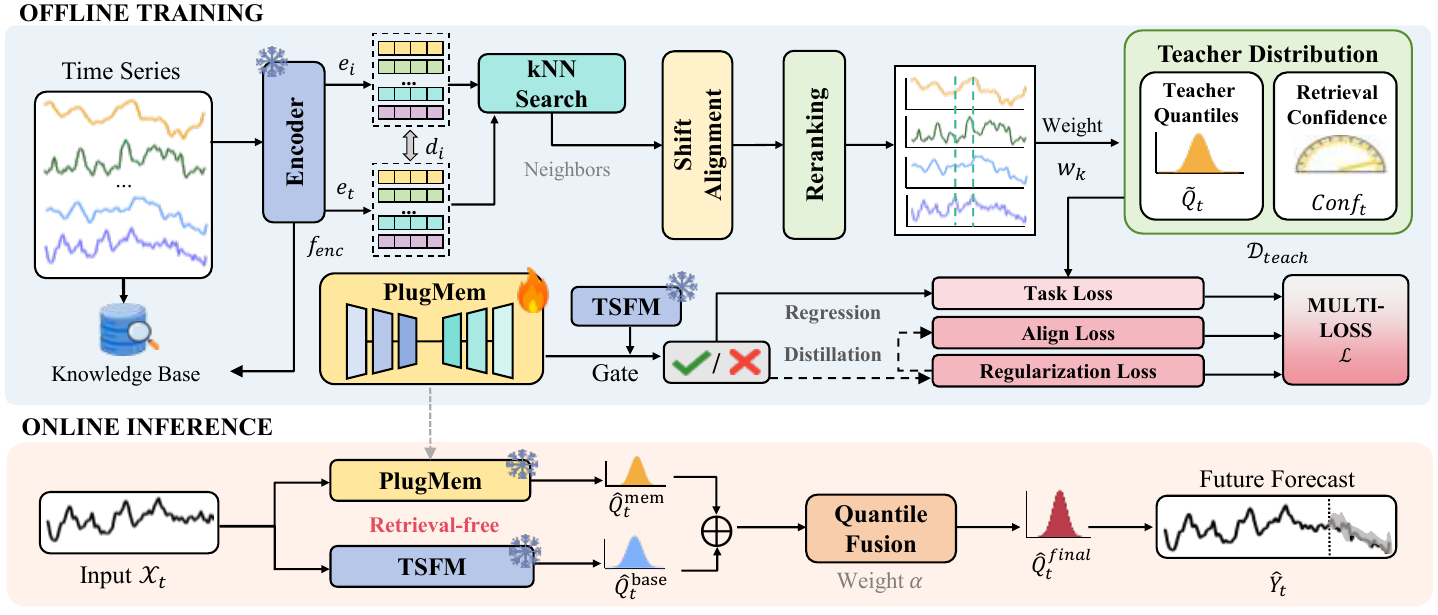}

  \caption{TS-Memory framework. }
  \Description{A two-phase pipeline diagram for TS-PlugMem. }
  \label{fig:ts_plugmem_overview}

\end{figure*}

\noindent\textbf{Adaptation Strategies for TSFMs.} 
To tackle domain shifts, existing work falls into two paradigms: \textit{(1) Parametric adaptation} updates model with target-domain data, ranging from full fine-tuning of all backbone weights, to partial fine-tuning of a small subset of parameters, and parameter-efficient add-ons that keep the backbone frozen while training lightweight modules such as adapters, LoRA-style low-rank updates, or prompt/prefix tuning~\cite{gupta2024low,gupta2024beyond,qiao2025multi,zhao2025less}. While these approaches reduce per-domain training cost, supporting many domains still incurs substantial maintenance overhead~\cite{ning2025mode,lee2025lightweight,sheng2023s}.
\textit{(2) Non-parametric retrieval} augments frozen models with external datastore evidence, with lightweight implementations relying on metric-based matching, where query contexts are compared to support embeddings via learned similarity functions for prediction~\cite{koch2015siamese,vinyals2016matching,snell2017prototypical,sung2018learning,oreshkin2018tadam}. For time series, it is commonly realized as subsequence retrieval (via DTW or other distance metrics) followed by aggregation over matched historical windows~\cite{tajmouati2024knn,martinez2019knn,martinez2018dealing,martinez2019tsfknn}. Inspired by NLP's RAG~\cite{lewis2020retrieval}, retrieval augmentation has been adapted to forecasting via relational retrieval, diffusion guidance, LLM-based retrieval and general frameworks~\cite{jing2022retrieval,liu2024retrieval,yang2025timerag,han2025retrieval}, with TSFM systems retrieving context-horizon pairs and fusing retrieved horizons for zero-shot forecasting~\cite{ning2025ts}. Despite their flexibility, these methods introduce heavy search operations into the critical inference path, leading to latency that scales linearly with database size~\cite{fan2024survey}.

We thus propose \textit{(3) Parametric Memory Distillation}. Inspired by memory-augmented networks~\cite{graves2014neural,weston2014memory} and NLP retrieval-free methods~\cite{tay2022transformer,izacard2021distilling}, we distill offline retrieval's distributional knowledge into a lightweight offline module, enabling TS-Memory to achieve retrieval-based adaptability while retaining the constant inference speed and single-model efficiency of frozen TSFMs.

\section{Problem Definition}
\label{sec:setup}


Let $\mathbf{x}_{1:T} \in \mathbb{R}^{T \times C}$ denote a multivariate time series with $C$ channels. At each time step $t$, the \emph{lookback context} $\mathbf{X}_t = \mathbf{x}_{t-L+1:t} \in \mathbb{R}^{L \times C}$ and \emph{forecast horizon} $\mathbf{Y}_t = \mathbf{x}_{t+1:t+H} \in \mathbb{R}^{H \times C}$ are defined accordingly. We target probabilistic forecasting via a set of quantile levels $\mathcal{Q}=\{q_k\}_{k=1}^Q$ with $q_k\in(0,1)$. For brevity, we use the multi-index $u=(h,c)$ where $u \in \mathcal{U} = \{1,\dots,H\} \times \{1,\dots,C\}$, so that $Y_{t,u}$ denotes the scalar value at time $t+h$ for channel $c$.

\noindent\textbf{Frozen Backbone.}
Let $f_{\theta}$ be a pre-trained TSFM with parameters $\theta$.
Given input $\mathbf{X}_t$, it outputs base quantile predictions:
\begin{equation}
    \widehat{\mathbf{Q}}^{\text{base}}_t = f_{\theta}(\mathbf{X}_t) \in \mathbb{R}^{Q \times H \times C}.
    \label{eq:tsm-base-quantile}
\end{equation}
Crucially, to preserve general knowledge and avoid storage overhead, $\theta$ remains frozen throughout the adaptation phase.

\noindent\textbf{Parametric Memory Module.}
We introduce \textbf{PlugMem}, a lightweight, learnable module $g_{\phi}$, parameterized by $\phi$.
It maps the same context $\mathbf{X}_t$ to a complementary set of quantile estimates:
\begin{equation}
    \widehat{\mathbf{Q}}^{\text{mem}}_t = g_{\phi}(\mathbf{X}_t) \in \mathbb{R}^{Q \times H \times C}.
    \label{eq:tsm-mem-quantile}
\end{equation}
Our objective is to optimize $\phi$ to distill retrieval-based distributional knowledge into $\widehat{\mathbf{Q}}^{\text{mem}}_t$, thereby achieving domain adaptation without accessing external databases during inference.

%% file: contents/3_method.tex
\section{Methodology}
\label{sec:method}


We propose \textbf{TS-Memory}, a lightweight retrieval-free memory module for adapting frozen TSFMs to target domains. While online retrieval effectively mitigates distribution shifts, it introduces heavy inference-time search overhead. Our core insight is that retrieval not only provides point estimates but also induces a non-parametric conditional distribution over plausible futures. Dispersion among retrieved neighbors inherently encodes context-dependent uncertainty, difficult to learn from a single realized trajectory. We treat this retrieval-induced distribution as privileged offline supervision and distill it into a parametric module, preserving retrieval-based adaptivity while enabling constant-time inference. Since constructing such targets is leakage-sensitive and too costly for the online path, we adopt a two-stage design as shown in Figure~\ref{fig:ts_plugmem_overview}.

\begin{itemize}[leftmargin=*]
    \item \textbf{Stage I: Privileged Supervision Construction.} To prevent test-time leakage, we construct a distributional teacher exclusively from the training set. We retrieve $K$ nearest neighbors in the frozen embedding space, synthesize their future trajectories into empirical quantiles, and thereby mine a privileged supervision signal ($\mathcal{D}_{\text{teach}}$) that encodes local domain patterns and uncertainty structures without relying on a run-time index.
    
    \item \textbf{Stage II: Confidence-Gated Memory Distillation.} We train the memory module to directly predict teacher quantiles from raw context. To mitigate noisy retrieval, we design a \textit{confidence-gated} dual objective: the module aligns with ground truth via standard regression, and distills the retrieval-induced distribution only when the teacher provides a reliable high-confidence improvement signal. At inference, PlugMem runs in parallel with the frozen backbone for robust constant-time forecasting.
\end{itemize}

\subsection{Privileged Supervision Construction}
\label{sec:teacher}

TS-Memory treats retrieval as training-time privileged supervision used only offline to build an auxiliary teacher dataset:
\begin{equation}
\mathcal{D}_{\text{teach}}
=\left\{(\mathbf{X}_t,\mathbf{Y}_t,\widetilde{\mathbf{Q}}_t,\mathrm{Conf}_t)\right\},
\label{eq:tsm-dteach}
\end{equation}
Concretely, a leakage-safe kNN teacher has privileged access to a knowledge base of past contexts and their realized futures, which is unavailable to the student at test time. This access allows the teacher to form a retrieval-conditioned predictive distribution, summarized as quantile targets $\widetilde{\mathbf{Q}}_t\in\mathbb{R}^{Q\times H\times C}$ with reliability score $\mathrm{Conf}_t\in[0,1]$. The construction process involves three steps:


\noindent\textbf{Knowledge Base Construction.} We first build a leakage-safe index utilizing the training split only: $\mathcal{K}=\{(\mathbf{X}^{(i)},\mathbf{Y}^{(i)})\}_{i=1}^{N_{\text{train}}}$,
ensuring that both the context window and forecast horizon are fully contained within the training segment. Then, we compute embeddings using a frozen encoder $f_{\text{enc}}$ (e.g., the encoder of the TSFM):
\begin{equation}
\mathbf{e}_t=f_{\text{enc}}(\mathbf{X}_t), \qquad\mathbf{e}_i=f_{\text{enc}}(\mathbf{X}^{(i)}).
\label{eq:tsm-emb}
\end{equation}
Candidates are retrieved via Euclidean distance $d_i=\|\mathbf{e}_t-\mathbf{e}_i\|_2$. We explicitly exclude the index-matched window to prevent trivial self-retrieval. Here, $\mathbf{e}_t$ is the embedding of the current input $\mathbf{X}_t$, and $\mathbf{e}_i$ is the embedding of the $i$-th candidate context. We use “retrieval-leakage-safe” to denote that all context–future pairs used by the teacher and retrieval baselines are constructed strictly from the training split; this does not assume that every TSFM pretraining corpus is fully auditable.

\noindent\textbf{Shift Alignment and Re-ranking.}
Distribution shift often manifests as level offsets, meaning that  neighbors close in embedding space may be mis-aligned in absolute scale. We align each candidate using a trailing mean shift over the last $m$ steps, computed channel-wise:
\begin{equation}
\boldsymbol{s}_i=\mathrm{mean}\!\left(\mathbf{X}_{t,L-m+1:L}\right)-\mathrm{mean}\!\left(\mathbf{X}^{(i)}_{L-m+1:L}\right)\in\mathbb{R}^{C}.
\label{eq:tsm-shift}
\end{equation}
This is applied to the context and future of the retrieved candidate:
\begin{equation}
\mathbf{X}^{(i)}_{\text{align}}=\mathbf{X}^{(i)}+\boldsymbol{s}_i,\qquad
\mathbf{Y}^{(i)}_{\text{align}}=\mathbf{Y}^{(i)}+\boldsymbol{s}_i.
\label{eq:tsm-align}
\end{equation}
Candidates are then re-ranked by the $\ell_1$ distance (i.e., $\text{score}_i=\|\mathbf{X}_t-\mathbf{X}^{(i)}_{\text{align}}\|_1$) between the query context $\mathbf{X}_t$ and the aligned candidate context $\mathbf{X}^{(i)}_{\text{align}}$, keeping the top $K$ neighbors.

\noindent\textbf{Teacher Aggregation and Confidence Scoring.}
Given the top $K$ neighbors with distances $\{d_k\}_{k=1}^{K}$, we compute softmax weights:
\begin{equation}
w_k=\frac{\exp\!\left(-\psi(d_k)/\tau_{\text{ret}}\right)}
{\sum_{j=1}^{K}\exp\!\left(-\psi(d_j)/\tau_{\text{ret}}\right)},
\qquad \sum_{k=1}^K w_k=1,
\label{eq:tsm-ret-weights}
\end{equation}
where $\psi(\cdot)$ is a monotone distance transform (identity by default) and $\tau_{\text{ret}}$ controls the weight sharpness. For each $u\in\mathcal{U}$, set $v_k=Y^{(k)}_{\text{align},u}$ and let $\pi$ sort $\{v_k\}$ in ascending order. Define cumulative weights $S_m=\sum_{r=1}^{m} w_{\pi(r)}$. Then, the weighted empirical quantile is:
\begin{equation}
\widetilde{Q}_{t,j,u}=v_{\pi(m^\star)},\qquad
m^\star=\min\{m:~S_m\ge q_j\}.
\label{eq:tsm-teacher-quantile}
\end{equation}
Here, $v_k$ is the aligned future value from the $k$-th neighbor, and $\pi(r)$ denotes the index of the $r$-th smallest value. We define retrieval confidence as the concentration of retrieval weights:
\begin{equation}
\mathrm{Conf}_t=\max_{1\le k\le K} w_k.
\label{eq:tsm-conf}
\end{equation}
The complete procedure is detailed in Algorithm~\ref{alg:teacher} (Appendix~\ref{appx:Offline_Teacher_Construction_Algorithm}).

\subsection{Confidence-Gated Memory Distillation}
\label{sec:plugmem}

We parameterize the memory module $g_{\phi}$ as a lightweight encoder--decoder Transformer. The input $\mathbf{X}_t$ is first normalized via Instance Normalization (Eq.~\ref{eq:tsm-inorm}), partitioned into patches of length $p$, and projected to $d$-dimensional tokens. A Transformer encoder produces a memory representation, which is attended to by $H$ learnable horizon queries in the decoder. A final quantile head maps these features to $Q$ quantile values, after which the normalization is inverted. Crucially, $g_{\phi}$ operates independently of the backbone internals, ensuring plug-and-play compatibility. 

Instance Normalization is defined as:
\begin{equation}
\widetilde{\mathbf{X}}_t = (\mathbf{X}_t - \mu(\mathbf{X}_t)) / (\sigma(\mathbf{X}_t) + \varepsilon).
\label{eq:tsm-inorm}
\end{equation}

To facilitate the memory module in learning transferable effective knowledge, we introduce three losses for joint optimization:

\noindent\textbf{Task Supervision $\mathcal{L}_{\text{task}}$.} Supervise PlugMem via standard quantile regression on ground-truth future values:
\begin{equation}
\mathcal{L}_{\text{task}} = \mathbb{E}_{t}\Big[
\operatorname{Pinball}_{\mathcal{Q}}\!\big(
\widehat{\mathbf{Q}}^{\text{mem}}_t,\mathbf{Y}_t
\big) \Big],
\label{eq:task-pinball-abstract}
\end{equation}
where $\operatorname{Pinball}_{\mathcal{Q}}(\cdot,\cdot)$ denotes the standard pinball loss for quantile regression, aggregated over the forecast horizon and variables.

\noindent\textbf{Confidence-Gated Distillation $\mathcal{L}_{\text{align}}$.} Distill the teacher's adaptive improvement into the memory module via confidence-gated distillation (two steps: error evaluation and incremental alignment), activated only when retrieval provides a reliable signal:

First, we evaluate the teacher's absolute prediction quality via median absolute error (median index $j^\star=\arg\min_j |q_j-0.5|$) to screen valid distillation windows, defining median absolute error for the retrieval teacher ($\mathrm{err}_t^T$) and frozen backbone ($\mathrm{err}_t^\mathrm{base}$) as:
\begin{equation}
\mathrm{err}_t^\star=\frac{1}{|\mathcal{U}|}\sum_{u\in\mathcal{U}}\bigl|Y_{t,u}-Q_{t,j^\star,u}^\star\bigr|,\;\star\in\{T,\mathrm{base}\},
\label{equ:err}
\end{equation}
where $\mathrm{err}_t^T$ and $\mathrm{err}_t^\mathrm{base}$ quantify the absolute deviation of the teacher and backbone predictions from the ground truth $Y_{t,u}$, respectively.

We then gate distillation with a margin $\epsilon_{\text{gate}}$ to retain only windows where the teacher outperforms the backbone:
\begin{equation}
\chi_t=\mathbb{I}\!\left(\mathrm{err}^{T}_t+\epsilon_{\text{gate}}<\mathrm{err}^{\text{base}}_t\right),\quad
\omega_t=\chi_t\cdot \mathrm{Conf}_t^{\gamma},
\label{eq:tsm-gate}
\end{equation}
where $\chi_t$ is the binary gating indicator (1 for valid windows), $\omega_t$ the confidence-weighted gating coefficient, $\mathrm{Conf}_t$ the retrieval confidence score, and $\gamma$ the confidence scaling hyperparameter.

\input{tables/different-TSFMs-with-TS-Memory-simple}

Second, we perform incremental alignment to distill the teacher's adaptive improvement (not global bias) into the memory module, aligning quantile predictions via the robust Huber loss $\ell_{\kappa}(\cdot,\cdot)$:
\begin{equation}
\mathcal{D}_{Q}(t) = \frac{1}{Q|\mathcal{U}|}\sum_{j=1}^{Q}\sum_{u\in\mathcal{U}}\ell_{\kappa}\!\left(\widehat{Q}^{\text{mem}}_{t,j,u},\widetilde{Q}_{t,j,u}\right).
\label{eq:tsm-dq}
\end{equation}
To mitigate sensitivity to global central bias of the frozen backbone, we further align the incremental median correction between the teacher and memory module (relative to the backbone):
\begin{equation}
\Delta^{\text{mem}}_{t,u}=\widehat{Q}^{\text{mem}}_{t,j^\star,u}-\widehat{Q}^{\text{base}}_{t,j^\star,u}, \quad
\qquad
\Delta^{T}_{t,u}=\widetilde{Q}_{t,j^\star,u}-\widehat{Q}^{\text{base}}_{t,j^\star,u},
\label{eq:tsm-delta}
\end{equation}
where $\Delta^{\text{mem}}_{t,u}$ and $\Delta^{T}_{t,u}$ denote the memory module and teacher's incremental median correction relative to the frozen backbone, respectively. The alignment loss for these corrections is:
\begin{equation}
\mathcal{D}_{\Delta}(t) = \frac{1}{|\mathcal{U}|}\sum_{u\in\mathcal{U}}
\ell_{\kappa}\!\left(\Delta^{\text{mem}}_{t,u},~\Delta^{T}_{t,u}\right).
\label{eq:tsm-ddelta}
\end{equation}
The final confidence-gated alignment loss is:
\begin{equation}
\mathcal{L}_{\text{align}}=\mathbb{E}_{t}\!\left[\omega_t\cdot\bigl(\mathcal{D}_{Q}(t)+\eta\,\mathcal{D}_{\Delta}(t)\bigr)\right].
\label{eq:tsm-loss-align}
\end{equation}

\noindent\textbf{Stability Regularization $\mathcal{L}_{\text{reg}}$.} Constrain the memory module to conservative behavior for uncertain retrieval (small $\omega_t$, weak $\mathcal{L}_{\text{align}}$ supervision), avoiding over-adjustment of the frozen backbone. We first anchor its median to the backbone with weight $(1-\omega_t)$:
\begin{equation}
\mathcal{L}_{\text{anchor}} = \mathbb{E}_{t}\!\left[(1-\omega_t)\cdot\frac{1}{|\mathcal{U}|}\sum_{u\in\mathcal{U}}\ell_{\kappa}\!\left(\widehat{Q}^{\text{mem}}_{t,j^\star,u},\widehat{Q}^{\text{base}}_{t,j^\star,u}\right)\right].
\label{eq:tsm-loss-anchor}
\end{equation}
We additionally impose a universal regularization to prevent quantile crossing in memory predictions, ensuring monotonic outputs:
\begin{equation}
\mathcal{L}_{\text{cross}} = \mathbb{E}_{t}\!\left[\frac{1}{(Q-1)|\mathcal{U}|}\sum_{u\in\mathcal{U}}\sum_{j=1}^{Q-1}\max\!\left(0,\widehat{Q}^{\text{mem}}_{t,j,u}-\widehat{Q}^{\text{mem}}_{t,j+1,u}\right)\right].
\label{eq:tsm-loss-cross}
\end{equation}
The final stability regularization loss is: $\mathcal{L}_{\text{reg}}=\mathcal{L}_{\text{anchor}}+\lambda_{\text{cross}}\,\mathcal{L}_{\text{cross}}$.

\noindent\textbf{Training objective.}
We train the memory module on $\mathcal{D}_{\text{teach}}$ using a composite loss that balances ground-truth regression with distributional distillation:
\begin{equation}
\mathcal{L} = \mathcal{L}_{\text{task}} + \lambda_{\text{align}}\,\mathcal{L}_{\text{align}} + \lambda_{\text{reg}}\,\mathcal{L}_{\text{reg}}.
\label{eq:tsm-loss-total}
\end{equation}

\subsection{Inference via Adaptive Fusion}
\label{sec:inference}

At test time, TS-Memory performs strictly retrieval-free inference:
\begin{equation}
\widehat{\mathbf{Q}}^{\text{base}}_t=f_{\theta}(\mathbf{X}_t),
\qquad
\widehat{\mathbf{Q}}^{\text{mem}}_t=g_{\phi}(\mathbf{X}_t).
\label{eq:tsm-infer}
\end{equation}
We fuse the two quantile forecasts by quantile-wise interpolation:
\begin{equation}
\widehat{\mathbf{Q}}^{\text{final}}_t
=(1-\alpha)\widehat{\mathbf{Q}}^{\text{base}}_t+\alpha\widehat{\mathbf{Q}}^{\text{mem}}_t,
\qquad
\alpha\in[0,1].
\label{eq:tsm-fuse}
\end{equation}
We tune $\alpha$ on a validation split.
Inference requires only two forward passes: one through the frozen backbone and one through the memory module.
No retrieval index needs to be maintained and no nearest-neighbor search is performed at serving time.

\noindent\textbf{Inference-time complexity.} TS-Memory adds constant overhead: one lightweight extra forward pass and quantile-wise fusion. Runtime and memory do not depend on the knowledge base since retrieval is offline. In contrast, online retrieval requires query-time search and aggregation that scale with the datastore.

%% file: tables/different-TSFMs-with-TS-Memory-simple.tex
\begin{table*}[!htbp]
\begin{center}
\caption{Long-term forecasting results of TS-Memory across frozen TSFM backbones. Results are averaged over forecasting horizons $H \in $\{96, 192, 336, 720\}. Lower values indicate better performance. Best results are highlighted in \textbf{bold}, and second
best results are underlined. Values in parentheses denote relative
improvement (\%). Full results are in Table~\ref{tab:ts_plugmem_full} of Appendix~\ref{appx:complete_results}.}

\begin{small}
\setlength\tabcolsep{3pt}
\renewcommand{\arraystretch}{0.85} 
\scalebox{0.9}{
\begin{tabular}{c|cc|cc|cc|cc|cc|cc|cc|cc}
\toprule
\multirow{1}{*}{TSFM} &
\multicolumn{4}{c|}{ChronosBolt} &
\multicolumn{4}{c|}{Chronos2} &
\multicolumn{4}{c|}{Sundial} &
\multicolumn{4}{c}{TimesFM} \\
\cmidrule{1-17}
\multirow{1}{*}{Dataset}
 &
\multicolumn{2}{c|}{Origin} & \multicolumn{2}{c|}{TS-Memory} &
\multicolumn{2}{c|}{Origin} & \multicolumn{2}{c|}{TS-Memory} &
\multicolumn{2}{c|}{Origin} & \multicolumn{2}{c|}{TS-Memory} &
\multicolumn{2}{c|}{Origin} & \multicolumn{2}{c}{TS-Memory} \\
\midrule
Metric &
MSE & MAE & MSE & MAE &
MSE & MAE & MSE & MAE &
MSE & MAE & MSE & MAE &
MSE & MAE & MSE & MAE \\
\midrule

ETTh1 & \underline{0.448} & \underline{0.419} & \textbf{0.421}\pct{-6.0\%} & \textbf{0.414}\pct{-1.2\%} & \underline{0.442} & \underline{0.412} & \textbf{0.420}\pct{-5.0\%} & \textbf{0.407}\pct{-1.2\%} & \underline{0.400} & \underline{0.412} & \textbf{0.396}\pct{-0.9\%} & \textbf{0.410}\pct{-0.5\%} & \underline{0.479} & \underline{0.442} & \textbf{0.448}\pct{-6.5\%} & \textbf{0.431}\pct{-2.5\%} \\
\midrule
ETTh2 & \underline{0.367} & \underline{0.380} & \textbf{0.354}\pct{-3.4\%} & \textbf{0.375}\pct{-1.1\%} & \underline{0.376} & \underline{0.383} & \textbf{0.350}\pct{-6.9\%} & \textbf{0.378}\pct{-1.5\%} & \underline{0.344} & \underline{0.380} & \textbf{0.340}\pct{-1.1\%} & \textbf{0.378}\pct{-0.6\%} & \underline{0.402} & \underline{0.409} & \textbf{0.364}\pct{-9.4\%} & \textbf{0.398}\pct{-2.9\%} \\
\midrule
ETTm1 & \underline{0.421} & \underline{0.383} & \textbf{0.381}\pct{-9.4\%} & \textbf{0.371}\pct{-3.1\%} & \underline{0.433} & \underline{0.381} & \textbf{0.393}\pct{-9.2\%} & \textbf{0.372}\pct{-2.3\%} & \underline{0.369} & \underline{0.369} & \textbf{0.356}\pct{-3.6\%} & \textbf{0.364}\pct{-1.5\%} & \underline{0.429} & \underline{0.416} & \textbf{0.382}\pct{-11.0\%} & \textbf{0.396}\pct{-4.8\%} \\
\midrule
ETTm2 & \underline{0.291} & \underline{0.317} & \textbf{0.267}\pct{-8.0\%} & \textbf{0.311}\pct{-2.1\%} & \underline{0.295} & \underline{0.315} & \textbf{0.270}\pct{-8.4\%} & \textbf{0.309}\pct{-2.0\%} & \underline{0.276} & \underline{0.317} & \textbf{0.266}\pct{-3.6\%} & \textbf{0.312}\pct{-1.6\%} & \underline{0.332} & \underline{0.341} & \textbf{0.279}\pct{-16.0\%} & \textbf{0.324}\pct{-5.0\%} \\
\midrule
Electricity & \underline{0.159} & \underline{0.244} & \textbf{0.155}\pct{-2.5\%} & \textbf{0.241}\pct{-1.0\%} & \underline{0.163} & \underline{0.244} & \textbf{0.159}\pct{-2.1\%} & \textbf{0.241}\pct{-1.1\%} & \underline{0.148} & \underline{0.242} & \textbf{0.144}\pct{-2.7\%} & \textbf{0.239}\pct{-1.4\%} & \underline{0.154} & \underline{0.244} & \textbf{0.149}\pct{-2.9\%} & \textbf{0.241}\pct{-1.1\%} \\
\midrule
Exchange & \underline{0.371} & \underline{0.412} & \textbf{0.365}\pct{-1.8\%} & \textbf{0.406}\pct{-1.5\%} & \underline{0.399} & \underline{0.421} & \textbf{0.366}\pct{-8.3\%} & \textbf{0.408}\pct{-3.1\%} & \underline{0.553} & \underline{0.494} & \textbf{0.471}\pct{-14.8\%} & \textbf{0.464}\pct{-6.0\%} & \underline{0.433} & \underline{0.446} & \textbf{0.425}\pct{-1.9\%} & \textbf{0.433}\pct{-2.7\%} \\
\midrule
Traffic & \underline{0.435} & \underline{0.263} & \textbf{0.425}\pct{-2.3\%} & \textbf{0.258}\pct{-1.6\%} & \underline{0.394} & \underline{0.237} & \textbf{0.389}\pct{-1.4\%} & \textbf{0.235}\pct{-0.6\%} & \underline{0.461} & \underline{0.286} & \textbf{0.436}\pct{-5.3\%} & \textbf{0.278}\pct{-3.0\%} & \underline{0.370} & \underline{0.244} & \textbf{0.367}\pct{-0.6\%} & \textbf{0.241}\pct{-1.1\%} \\
\midrule
Weather & \underline{0.263} & \underline{0.276} & \textbf{0.239}\pct{-9.1\%} & \textbf{0.267}\pct{-3.5\%} & \underline{0.274} & \underline{0.267} & \textbf{0.241}\pct{-12.0\%} & \textbf{0.260}\pct{-2.6\%} & \underline{0.244} & \underline{0.269} & \textbf{0.238}\pct{-2.5\%} & \textbf{0.266}\pct{-1.2\%} & \underline{0.222} & \underline{0.237} & \textbf{0.205}\pct{-7.6\%} & \textbf{0.230}\pct{-2.9\%} \\

\bottomrule
\end{tabular}}
\end{small}
\label{tab:ts_plugmem_compact}
\end{center}

\end{table*}

%% file: contents/4_experiment.tex
\section{Experiment}
\label{sec:experiments}

We evaluate TS-Memory on long-horizon multivariate forecasting with four research questions (\textbf{RQs}):

\begin{itemize}[leftmargin=*]
    \item \textbf{RQ1:} Can TS-Memory consistently improve various TSFMs across diverse datasets and different backbones? $\rightarrow$ Sec.~\ref{sec:ts_memory_perf_backbones}
    \item \textbf{RQ2:} How does TS-Memory compare with other adaptation strategies in performance and inference latency? $\rightarrow$ Sec.~\ref{sec:comparison_adaptation_baselines}
    \item \textbf{RQ3:} How does TS-Memory perform under backbone scaling and cross-model/cross-dataset transfer? $\rightarrow$ Sec.~\ref{sec:transfer_generality}
    \item \textbf{RQ4:} What are the key design factors of TS-Memory, how do parameter scaling and sensitivity impact its performance, and what qualitative results does it achieve? $\rightarrow$ Sec.~\ref{sec:model_analysis} 
\end{itemize}

\input{tables/ragvsTS-Memory-simple}
\input{tables/efficiency_query-simple}

\subsection{Experimental Setup}

\noindent\textbf{Datasets \& Evaluation Metrics.} We experiment on eight widely used multivariate long-term forecasting benchmarks (ETTh1/2, ETTm1/2, Electricity, Exchange-rate, Traffic, Weather)~\cite{zhou2021informer,lai2018modeling}. Following the standard long-horizon protocol, we evaluate across multiple forecasting horizons and report the average performance to characterize overall long-range forecasting ability~\cite{wu2023timesnet}. For point forecasting, we report \textit{MSE} and \textit{MAE}~\cite{hyndman2006another}; for probabilistic forecasting, we additionally present \textit{CRPS} to assess prediction quality~\cite{gneiting2007strictly}.

\noindent\textbf{Baselines \& Backbones.} 
We adopt four representative TSFMs: \textit{ChronosBolt}~\cite{ansari2024chronos}, \textit{Chronos2}~\cite{ansari2025chronos}, \textit{Sundial}~\cite{liu2025sundial} and \textit{TimesFM}~\cite{das2024decoder}, with backbone parameters frozen unless specified otherwise. 
We compare against three baselines: 
\textit{(1)} \textit{Origin}, the zero-shot deployment of frozen backbones; 
\textit{(2)} Online retrieval augmentation baselines (\textit{RAFT}~\cite{han2025retrieval}, \textit{TS-RAG}~
\cite{ning2025ts}), which perform test-time $k$NN retrieval from a leakage-safe datastore and fuse retrieved futures for forecasting; 
\textit{(3)} The parameter-efficient fine-tuning baseline \textit{LoRA}, which injects low-rank adapters into attention projection layers and only trains the adapters, with trainable parameters matched to PlugMem.

\noindent\textbf{Implementation Details.} Following prior work~\cite{wu2023timesnet}, we use prediction horizons $H \in \{96, 192, 336, 720\}$ and a fixed look-back window of $L = 512$. All models are optimized via Adam~\cite{kingma2014adam}. For baseline methods on backbone models, we strictly adhere to their original hyperparameter settings to ensure fair comparison. All experiments are implemented in PyTorch and run on NVIDIA A800 80GB GPUs. Additional details are provided in Appendix~\ref{appx:experiment_details}.

\subsection{TS-Memory Performance Across Backbones}
\label{sec:ts_memory_perf_backbones}

Table~\ref{tab:ts_plugmem_compact} presents long-horizon forecasting results for TS-Memory deployed on four frozen TSFM backbones across eight benchmarks. TS-Memory enhances all backbones on every dataset, confirming that retrieval-driven gains can be distilled into a compact memory module without backbone parameter update. Averaged across all dataset–backbone pairs, TS-Memory reduces MSE by 5.8\% and MAE by 2.1\%, with peak reductions of 16.0\% (MSE) and 6.0\% (MAE). Gains are most pronounced on benchmarks with severe distribution shifts and rich temporal structures (e.g., Weather and ETT minute-level datasets), where double-digit MSE reductions are common, demonstrating improved robustness to long-horizon error accumulation. For regular benchmarks (e.g., Electricity), improvements are modest yet consistent. It matches the desired plug-and-play adapter behavior: substantial gains for poorly performing zero-shot backbones, and stability for well-aligned ones. Consistent MAE reductions further confirm broad error mitigation across time steps and variables, rather than sporadic gains in isolated cases.

\noindent\textbf{Pretraining-overlap check.}
Our retrieval-leakage-safe construction rules out test-window leakage for the offline teacher and online retrieval baselines, but does not assume fully auditable TSFM pretraining corpora. We therefore exclude confirmable backbone--dataset overlap pairs: TimesFM on Electricity, Traffic, and Weather; Sundial on Traffic; and Chronos2 on Electricity. On the $27$ pairs, TS-Memory reduces MSE by $6.22\%$ and MAE by $2.18\%$, slightly stronger than the full $32$-pair reductions of $5.8\%$ and $2.1\%$. This suggests our gains are not driven by known pretraining overlap.

\input{tables/different-size-of-Chronos-simple}

\input{tables/plugmem_merged_table-simple}

\subsection{Comparison with Adaptation Baselines}
\label{sec:comparison_adaptation_baselines}


\noindent\textbf{Performance comparison.}
Table~\ref{tab:chronosbolt_base_ablation_compact} compares TS-Memory with online retrieval baselines (RAFT and TS-RAG) and the parameter-efficient fine-tuning baseline LoRA on ChronosBolt, reporting point (MSE/MAE) and probabilistic (CRPS) metrics. Our method is the only method that consistently reduces MSE, MAE, and CRPS across all eight datasets, indicating that retrieval-distilled corrections can improve not only the mean prediction but also the reliability of the predictive distribution. In contrast, LoRA yields competitive reductions in MSE/MAE on some datasets, though its CRPS behavior is inconsistent and can even degrade on certain benchmarks, suggesting that low-rank parameter update may overfit point objectives or disturb the model's uncertainty calibration. Overall, these results support the core motivation of TS-Memory: distilling retrieval guidance into a lightweight module can preserve the retrieval style robustness while maintaining stable probabilistic forecasts.

\noindent\textbf{Inference latency comparison.}
Table~\ref{tab:chronosbolt_time_latency_compact} reports inference latency for all methods under the same evaluation protocol. Online retrieval introduces substantial query-dependent overhead: TS-RAG leads to nearly two-fold end-to-end slowdown, with retrieval dominating total latency, while RAFT is more efficient but still incurs noticeable overhead, especially when the datastore grows. LoRA avoids retrieval during inference, but still adds non-trivial constant latency due to layer-wise low-rank projections. TS-Memory removes retrieval entirely and introduces only a small constant overhead via a single lightweight forward pass, making its runtime close to the frozen-backbone baseline. This efficiency profile is important in deployment settings where retrieval infrastructure can become the dominant bottleneck due to index maintenance costs, latency variance, and poor scaling with horizon or datastore size.

\subsection{Transfer generality}
\label{sec:transfer_generality}

\noindent\textbf{Scaling across model sizes.}
\label{sec:transfer_scaling_model_sizes}
Table~\ref{tab:chronosbolt_ts_memory_compact} shows that a TS-Memory module trained once on ChronosBolt-Base (205M) can be reused as-is on smaller ChronosBolt variants (down to 9M parameters). Across all eight benchmarks, it consistently improves MSE/MAE without re-optimization, implying that the distilled correction signal is not tied to a specific backbone capacity. The average MSE gain remains around 5\%, and improvements are often larger on challenging datasets (e.g., Weather and minute-level ETT), where smaller frozen backbones are more prone to long-horizon drift. It suggests that TS-Memory can act as a constant-cost adapter that upgrades compute-constrained backbones without retrieval infrastructure.

\input{tables/Train-test-domain-configurations}

\noindent\textbf{Cross-backbone transfer.}
\label{sec:transfer_cross_backbone}
Table~\ref{tab:plugmem_merged_compact} evaluates transfer across TSFM families by distilling PlugMem from a single retrieval teacher and applying it to different frozen target backbones. We observe strong cross-model transfer: PlugMem distilled from Chronos2 improves ChronosBolt, Sundial, and TimesFM, and PlugMem distilled from Sundial similarly benefits ChronosBolt, Chronos2, and TimesFM. Average MSE reductions fall in the 6\%--10\% range depending on the target backbone, supporting that TS-Memory captures generalizable retrieval-driven correction patterns. MAE gains are sometimes smaller and can be near neutral for some teacher--target pairs, which is expected because different architectures may exhibit different output calibration and loss sensitivity. However, the overall trend remains consistently positive across targets. 

Cross-backbone transfer is enabled by three design choices. First, PlugMem consumes the raw context window rather than backbone hidden states, making it independent of TSFM internals. Second, it predicts quantiles in the common forecasting output space, so the learned memory can be fused with different probabilistic backbones. Third, the $\Delta$-alignment objective distills retrieval-induced corrections relative to the frozen backbone, encouraging PlugMem to learn domain-level temporal correction patterns rather than model-specific representations.


\begin{figure}[!t]
  \centering
  \includegraphics[width=\linewidth]{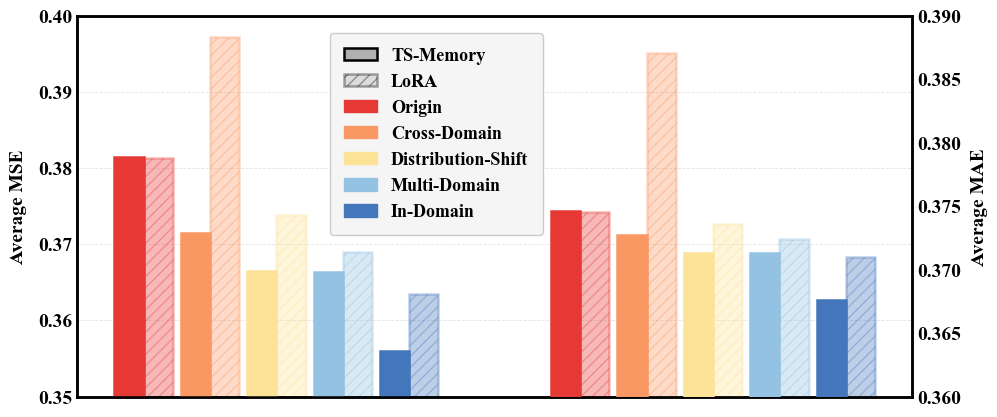}
  \caption{TS-Memory vs. LoRA under different train-test domains. Full per-dataset results are provided in Table~\ref{tab:domain_split_lora_tsmemory}.}
  \Description{TS-Memory vs. LoRA under different train-test domains.}
  \label{tab:domain_split_lora_tsmemory_bar}
\end{figure}

\begin{figure}[!b]
  \centering
  \includegraphics[width=\linewidth]{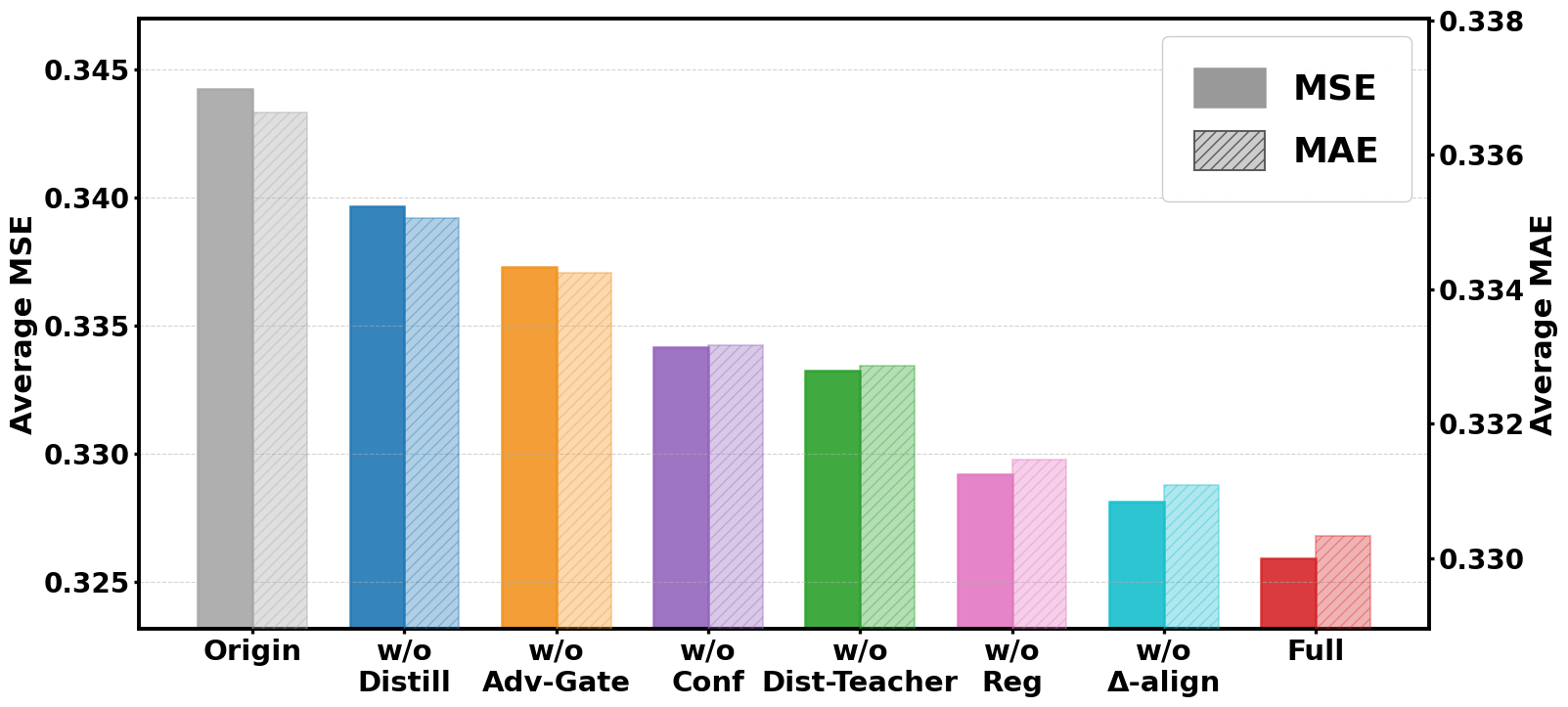}
  \caption{Ablation study of TS-Memory components.}
  \Description{Ablation study of TS-Memory components.}
  \label{fig:ablation_study}
\end{figure}

\noindent\textbf{Generalization under domain shift.}
\label{sec:transfer_domain_shift}
Figure~\ref{tab:domain_split_lora_tsmemory_bar} evaluates our method under domain shift on the ChronosBolt backbone. 
Table~\ref{tab:setting_test_relation} summarizes the train$\rightarrow
$test domain relations, where \textit{train} denotes the domain used to construct offline retrieval supervision and \textit{test} the target domain for evaluation.
Even with mismatched supervision, TS-Memory provides non-trivial gains: cross-domain training reduces average MSE by 2.7\%, while more related transfer settings yield larger and more stable improvements, reaching 3.5\% MSE reduction under distribution-shift transfer and 4.0\% under multi-domain transfer. In-domain supervision performs best, achieving 6.8\% lower MSE and 1.9\% lower MAE on average, confirming that retrieval supervision is most effective when the knowledge base aligns with the target distribution. This smooth scaling with alignment suggests that TS-Memory does not rely on brittle dataset-specific shortcuts, but instead distills transferable temporal motifs. Meanwhile, LoRA exhibits weaker cross-domain robustness and can underperform the frozen backbone under domain mismatch.

\subsection{Model Analysis}
\label{sec:model_analysis}

\noindent\textbf{Ablation study.}
\label{sec:model_analysis_ablation}
Figure~\ref{fig:ablation_study} reports component ablations averaged across datasets and horizons, where the full TS-Memory achieves the best overall performance (5.3\% lower MSE and 1.9\% lower MAE compared to the frozen backbone). When retrieval distillation is removed, the gains largely vanish, indicating that the offline retrieval teacher is the primary source of transferable corrective signal. Moreover, disabling either the advantage gate or the confidence-aware weighting consistently degrades performance, suggesting that \textit{selective} and \textit{confidence-conditioned} distillation is crucial: it filters unreliable retrieval targets and prevents negative transfer when retrieved futures are noisy or mismatched. Finally, auxiliary stability terms (e.g., Reg and $\Delta$-align) yield modest yet consistent improvements, primarily by stabilizing training and discouraging over-correction so that the learned memory behaves as a conservative correction module on top of the strong frozen prior.

\begin{figure}[!t]
  \centering
  \includegraphics[width=\linewidth]{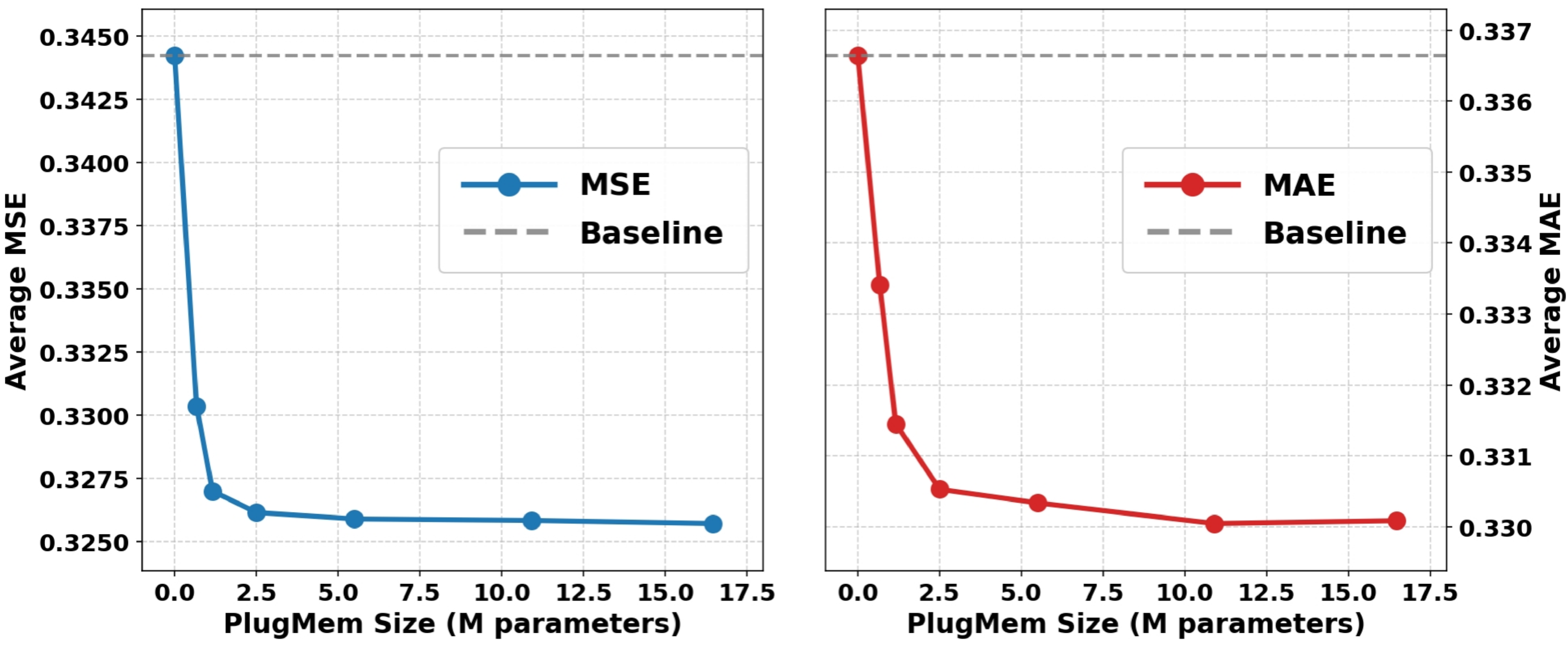}

  \caption{Scaling Analysis of PlugMem Capacity. }
  \Description{Scaling analysis of PlugMem for long-term forecasting.}
  \label{fig:Scaling Analysis of PlugMem}

\end{figure}

\noindent\textbf{Scaling PlugMem capacity.}
\label{sec:model_analysis_scaling_capacity}
Figure~\ref{fig:Scaling Analysis of PlugMem} studies the effect of PlugMem capacity under a fixed training recipe and frozen backbone, focusing on long-horizon forecasting. Even a small PlugMem recovers most of the gain over the frozen TSFM, suggesting that the retrieval teacher mainly provides low-complexity correction patterns rather than requiring large task-specific parameterization. As capacity increases beyond a moderate size, improvements saturate and can slightly regress on some datasets, indicating over-correction or idiosyncratic artifacts from the teacher. Overall, a moderately sized PlugMem offers the best accuracy–efficiency trade-off for plug-and-play deployment. Accordingly, we design three PlugMem variants of different sizes (see Appendix~\ref{appx:complete_results}, Table~\ref{tab:mem-size}).

\begin{figure}[!b]
  \centering

  \includegraphics[width=\linewidth]{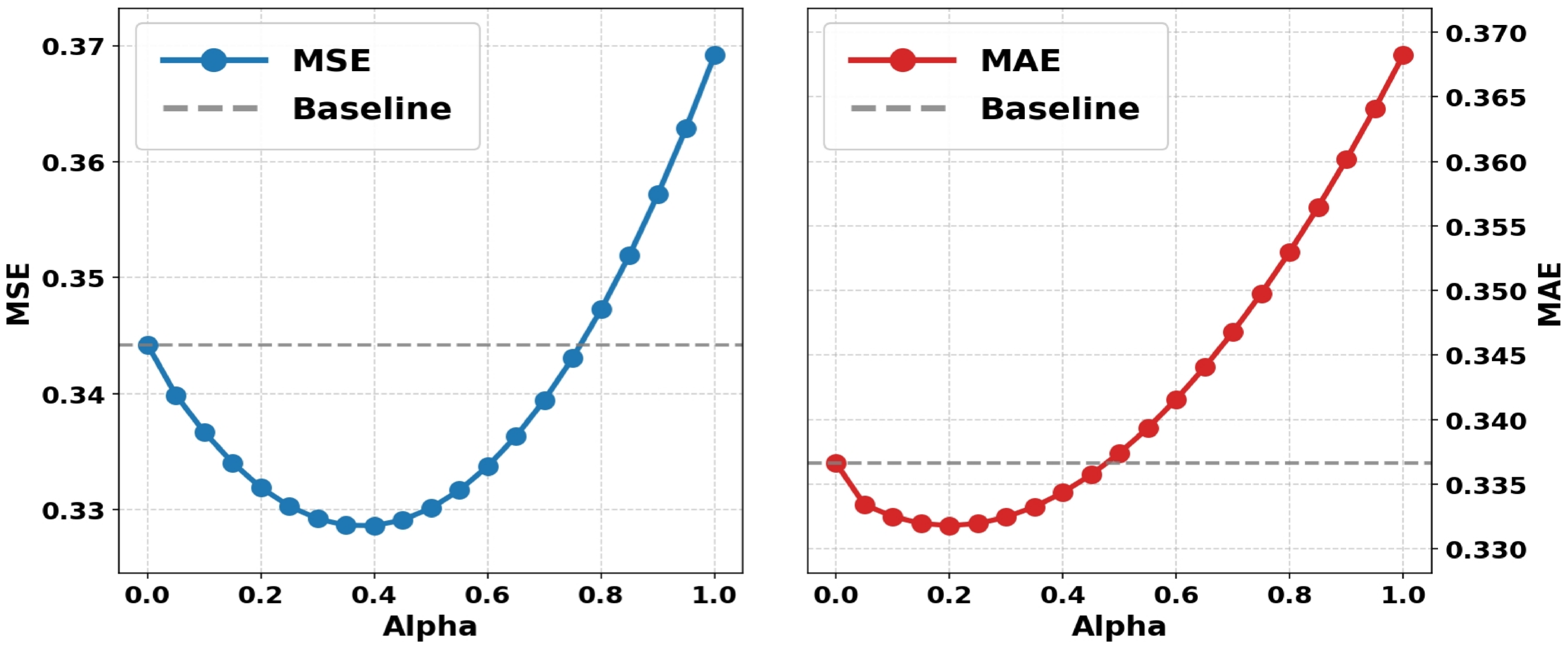}

  \caption{Sensitivity Analysis of the Fusion Weight $\alpha$.}
  \Description{Sensitivity analysis of $\alpha$ for long-term forecasting.}
  \label{fig:sensitivity_analysis}
\end{figure}

\noindent\textbf{Sensitivity to the fusion weight.}
\label{sec:model_analysis_sensitivity_alpha}
Figure~\ref {fig:sensitivity_analysis} analyzes the fusion weight $\alpha$ for combining backbone and PlugMem quantile forecasts. Performance improves as $\alpha$ increases from 0, confirming that the memory branch provides complementary corrections rather than redundant predictions. However, excessively large $\alpha$ degrades MSE/MAE, as over-reliance on memory overrides the strong priors learned by the backbone and amplifies residual bias; in the extreme case of $\alpha=1$, the backbone is discarded entirely in favor of PlugMem alone. Notably, the curve exhibits an optimum and remains flat near it, enabling stable, lightweight validation-based tuning.

\noindent\textbf{Gate statistics.}
Table~\ref{tab:gate_stats} summarizes how the advantage gate behaves during memory distillation. 
Here, $\chi_t{=}1$ indicates that the retrieval teacher outperforms the frozen backbone on a training window, $\omega_t$ is the confidence-weighted gate coefficient, and Adv. Margin measures the teacher's improvement over the backbone. 
Across all eight datasets, the teacher outperforms the backbone on $83.5\%$ of training windows, but the gate is not always-on. 
Activation and $\omega_t$ are lower on well-calibrated benchmarks such as Electricity and Traffic, while activation is much higher on shifting domains such as Weather and Exchange. 
This indicates that the gate acts as a domain-adaptive filter, selectively distilling useful retrieval signals while suppressing weak or uncertain teacher corrections.

\begin{table}[t]
\centering
\small
\caption{Gate activation statistics during confidence-gated memory distillation.}
\label{tab:gate_stats}
\begin{tabular*}{\linewidth}{@{\extracolsep{\fill}}lcccc}
\toprule
Dataset & Act. $(\chi_t{=}1)$ & Mean $\omega_t$ & Mean Conf. & Adv. Margin \\
\midrule
ETT Avg     & $87.7\%$ & 0.479 & 0.562 & 0.173 \\
Weather     & $96.7\%$ & 0.680 & 0.701 & 0.245 \\
Traffic     & $68.1\%$ & 0.067 & 0.100 & 0.025 \\
Electricity & $50.8\%$ & 0.062 & 0.113 & 0.013 \\
Exchange    & $99.4\%$ & 0.102 & 0.102 & 0.184 \\
\bottomrule
\end{tabular*}
\end{table}

\begin{figure}[!b]
  \centering
  \includegraphics[width=\linewidth]{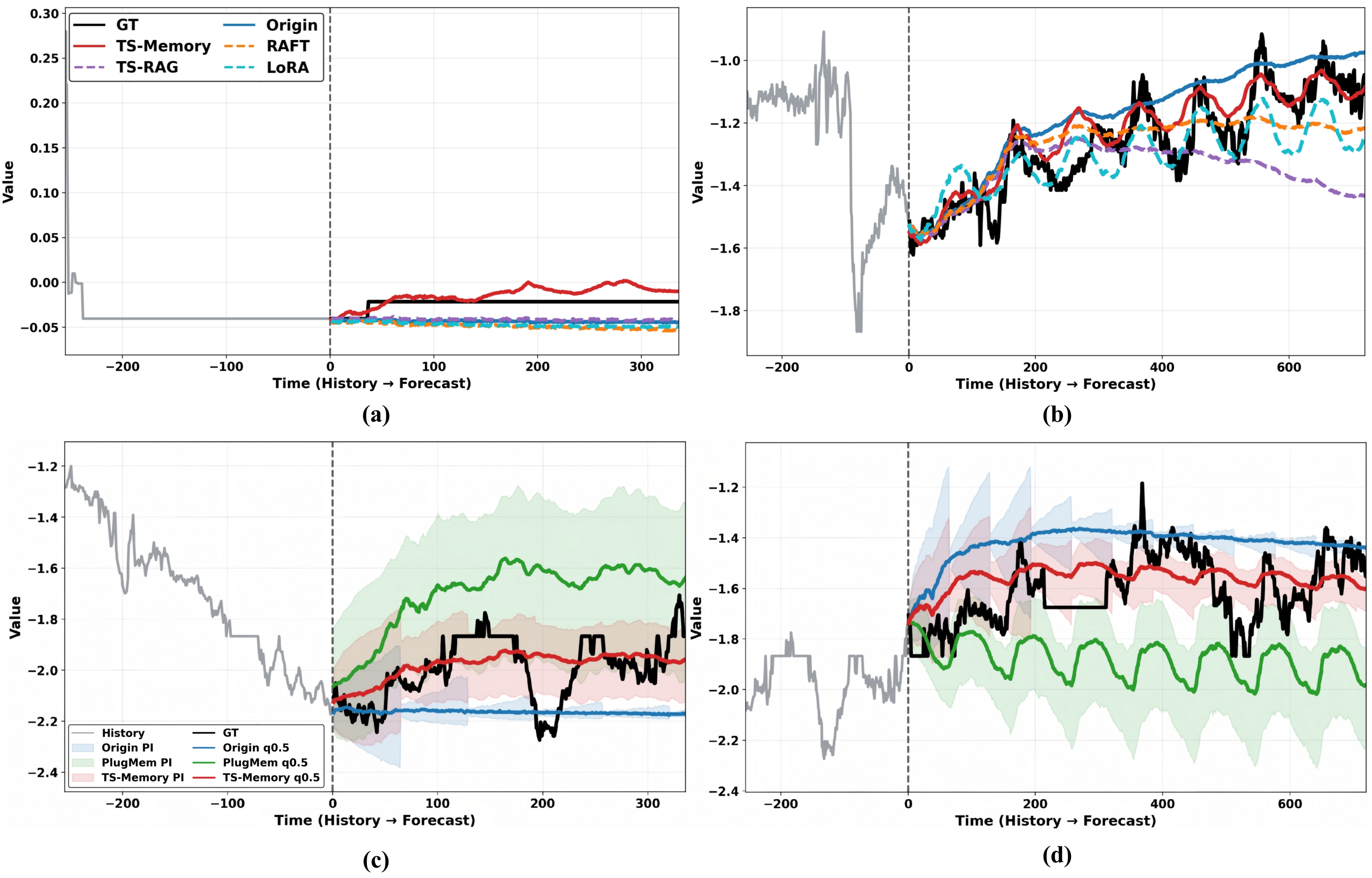}

  \caption{Qualitative visualizations comparing TS-Memory with the frozen backbone and adaptation baselines.}
  \Description{Qualitative Visualization.}
  \label{fig:qualitative_visualization}

\end{figure}
\noindent\textbf{Qualitative visualizations.}
Figure~\ref{fig:qualitative_visualization} presents representative cases highlighting behavioral differences among methods. In (a), the frozen backbone and several baselines remain nearly flat after the forecast start (dashed line), failing to track the target's subsequent level shift, whereas TS-Memory produces a clear corrective trend that is better aligned with the ground truth, consistent with learning compact bias corrections from retrieval distillation. In (b), TS-RAG exhibits noticeable long-horizon drift, and LoRA/RAFT introduce phase or smoothness artifacts, while TS-Memory stays closer to the ground-truth trajectory with reduced accumulated error, demonstrating that distilling retrieval offline into a memory module preserves adaptability without the instability inherent in online retrieval. In (c,d), probabilistic forecast comparisons illustrate why selective, confidence-aware training matters: PlugMem alone can become overly aggressive, producing a mis-centered median and excessively wide or oscillatory prediction intervals, while the frozen backbone tends to be conservative but biased; TS-Memory balances both, yielding well-calibrated interval width and a more accurately centered median. Overall, these plots corroborate the ablation findings: advantage-gated, confidence-weighted distillation enables TS-Memory to absorb useful retrieval signals while suppressing unreliable teacher artifacts, producing a robust correction mechanism rather than a fragile replacement predictor.

%% file: tables/ragvsTS-Memory-simple.tex
\begin{table*}[!htbp]
\begin{center}
\caption{Long-term forecasting results comparing TS-Memory with online retrieval baselines on ChronosBolt. We use the same protocol as in Table~\ref{tab:ts_plugmem_compact}. Full results are in Table~\ref{tab:chronosbolt_base_ablation} of Appendix~\ref{appx:complete_results} .}
\setlength\tabcolsep{3pt}
\renewcommand{\arraystretch}{0.80} 
\scalebox{0.80}{
\begin{tabular}{c|ccc|ccc|ccc|ccc|ccc}
\toprule
Dataset &
\multicolumn{3}{c|}{Origin} &
\multicolumn{3}{c|}{RAFT} &
\multicolumn{3}{c|}{TS-RAG} &
\multicolumn{3}{c|}{LoRA} &
\multicolumn{3}{c}{TS-Memory} \\
\midrule
Metric &
MSE & MAE & CRPS &
MSE & MAE & CRPS &
MSE & MAE & CRPS &
MSE & MAE & CRPS &
MSE & MAE & CRPS \\
\midrule
ETTh1 & 0.448 & 0.419 & 0.437 & 0.437\pct{-2.5\%} & 0.418\pct{-0.2\%} & 0.432\pct{-1.1\%} & 0.442\pct{-1.3\%} & 0.421\pct{+0.7\%} & 0.440\pct{+0.8\%} & \underline{0.428}\pct{-4.4\%} & \underline{0.416}\pct{-0.6\%} & \underline{0.428}\pct{-2.0\%} & \textbf{0.421}\pct{-6.0\%} & \textbf{0.414}\pct{-1.2\%} & \textbf{0.407}\pct{-6.8\%} \\
\midrule
ETTh2 & 0.367 & 0.380 & 0.246 & 0.357\pct{-2.5\%} & 0.378\pct{-0.5\%} & \underline{0.243}\pct{-1.0\%} & 0.362\pct{-1.4\%} & 0.385\pct{+1.5\%} & 0.250\pct{+1.6\%} & \underline{0.357}\pct{-2.7\%} & \underline{0.377}\pct{-0.6\%} & 0.252\pct{+2.6\%} & \textbf{0.354}\pct{-3.4\%} & \textbf{0.375}\pct{-1.1\%} & \textbf{0.238}\pct{-3.1\%} \\
\midrule
ETTm1 & 0.421 & 0.383 & 0.409 & 0.398\pct{-5.5\%} & 0.377\pct{-1.6\%} & 0.400\pct{-2.3\%} & 0.406\pct{-3.6\%} & 0.381\pct{-0.4\%} & 0.406\pct{-0.9\%} & \underline{0.392}\pct{-6.9\%} & \underline{0.376}\pct{-1.7\%} & \underline{0.397}\pct{-2.9\%} & \textbf{0.381}\pct{-9.4\%} & \textbf{0.371}\pct{-3.1\%} & \textbf{0.370}\pct{-9.7\%} \\
\midrule
ETTm2 & 0.291 & 0.317 & 0.205 & 0.281\pct{-3.3\%} & 0.315\pct{-0.9\%} & 0.203\pct{-1.2\%} & 0.283\pct{-2.6\%} & 0.318\pct{+0.3\%} & 0.205\pct{-0.2\%} & \underline{0.278}\pct{-4.5\%} & \underline{0.314}\pct{-1.2\%} & \underline{0.199}\pct{-3.2\%} & \textbf{0.267}\pct{-8.0\%} & \textbf{0.311}\pct{-2.1\%} & \textbf{0.190}\pct{-7.6\%} \\
\midrule
Electricity & 0.159 & 0.244 & \underline{0.239} & 0.156\pct{-1.5\%} & 0.244\pct{+0.2\%} & 0.240\pct{+0.2\%} & 0.157\pct{-1.4\%} & 0.245\pct{+0.2\%} & 0.240\pct{+0.3\%} & \underline{0.156}\pct{-1.8\%} & \underline{0.244}\pct{-0.1\%} & 0.244\pct{+1.9\%} & \textbf{0.155}\pct{-2.5\%} & \textbf{0.241}\pct{-1.0\%} & \textbf{0.232}\pct{-3.2\%} \\
\midrule
Exchange & 0.371 & 0.412 & 0.268 & 0.369\pct{-0.7\%} & 0.409\pct{-0.9\%} & 0.257\pct{-4.2\%} & 0.367\pct{-1.2\%} & 0.408\pct{-1.0\%} & \underline{0.253}\pct{-5.6\%} & \underline{0.366}\pct{-1.4\%} & \underline{0.407}\pct{-1.2\%} & 0.266\pct{-0.8\%} & \textbf{0.365}\pct{-1.8\%} & \textbf{0.406}\pct{-1.5\%} & \textbf{0.250}\pct{-6.9\%} \\
\midrule
Traffic & 0.435 & 0.263 & 0.273 & 0.426\pct{-2.0\%} & 0.261\pct{-0.8\%} & 0.271\pct{-0.9\%} & 0.427\pct{-1.9\%} & 0.261\pct{-0.7\%} & 0.271\pct{-0.8\%} & \underline{0.425}\pct{-2.2\%} & \underline{0.260}\pct{-1.1\%} & \underline{0.268}\pct{-2.0\%} & \textbf{0.425}\pct{-2.3\%} & \textbf{0.258}\pct{-1.6\%} & \textbf{0.267}\pct{-2.4\%} \\
\midrule
Weather & 0.263 & 0.276 & 0.404 & 0.254\pct{-3.5\%} & \underline{0.268}\pct{-3.0\%} & 0.390\pct{-3.5\%} & 0.262\pct{-0.6\%} & 0.279\pct{+1.2\%} & 0.408\pct{+1.0\%} & \underline{0.247}\pct{-6.1\%} & 0.268\pct{-2.8\%} & \underline{0.378}\pct{-6.4\%} & \textbf{0.239}\pct{-9.1\%} & \textbf{0.267}\pct{-3.5\%} & \textbf{0.370}\pct{-8.4\%} \\
\bottomrule
\end{tabular}}
\label{tab:chronosbolt_base_ablation_compact}
\end{center}

\end{table*}

%% file: tables/efficiency_query-simple.tex
\begin{table*}[!htbp]
\begin{center}
\caption{Inference latency comparison (ms/query) for Origin, online retrieval baselines, LoRA, and TS-Memory on ChronosBolt. We report retrieval time (Retr), forward time (Fwd), total time (Total), and retrieval fraction (Frac), following the same protocol as Table~\ref{tab:ts_plugmem_compact}. Full results are in  Table~\ref{tab:chronosbolt_time_latency} of Appendix~\ref{appx:complete_results}.}

\renewcommand{\arraystretch}{0.80} 
\setlength\tabcolsep{2pt}
\scalebox{0.83}{
\begin{tabular}{c|cccc|cccc|cccc|cccc|cccc}
\toprule
\multirow{1}{*}{Method} &
\multicolumn{4}{c|}{Origin} &
\multicolumn{4}{c|}{RAFT} &
\multicolumn{4}{c|}{TS-RAG} &
\multicolumn{4}{c|}{LoRA} &
\multicolumn{4}{c}{TS-Memory} \\
\midrule
\multirow{1}{*}{Metric}
& Retr & Fwd & Total & Frac &
Retr & Fwd & Total & Frac &
Retr & Fwd & Total & Frac &
Retr & Fwd & Total & Frac &
Retr & Fwd & Total & Frac \\
\midrule
ETTh1
& 0 & 3.53 & 3.53 & 0\%
& 0.84 & 3.84\pct{+8.8\%} & 4.68\pct{+32.6\%} & 20.0\%
& 3.53 & \underline{3.80}\pct{+7.8\%} & 7.33\pct{+107.9\%} & 51.1\%
& \underline{0} & 4.28\pct{+21.4\%} & \underline{4.28}\pct{+21.4\%} & \underline{0\%}
& \textbf{0} & \textbf{3.69}\pct{+4.7\%} & \textbf{3.69}\pct{+4.7\%} & \textbf{0\%} \\
\midrule
ETTh2
& 0 & 3.52 & 3.52 & 0\%
& 0.84 & \underline{3.82}\pct{+8.4\%} & 4.65\pct{+32.1\%} & 19.7\%
& 3.50 & 3.83\pct{+8.7\%} & 7.33\pct{+108.2\%} & 50.6\%
& \underline{0} & 4.29\pct{+21.7\%} & \underline{4.29}\pct{+21.7\%} & \underline{0\%}
& \textbf{0} & \textbf{3.66}\pct{+4.0\%} & \textbf{3.66}\pct{+4.0\%} & \textbf{0\%} \\
\midrule
ETTm1
& 0 & 3.52 & 3.52 & 0\%
& 2.62 & 3.79\pct{+7.5\%} & 6.41\pct{+81.9\%} & 43.9\%
& 6.46 & \underline{3.78}\pct{+7.4\%} & 10.24\pct{+190.6\%} & 65.7\%
& \underline{0} & 4.24\pct{+20.4\%} & \underline{4.24}\pct{+20.4\%} & \underline{0\%}
& \textbf{0} & \textbf{3.67}\pct{+4.1\%} & \textbf{3.67}\pct{+4.1\%} & \textbf{0\%} \\
\midrule
ETTm2
& 0 & 3.52 & 3.52 & 0\%
& 2.52 & \underline{3.79}\pct{+7.5\%} & 6.30\pct{+79.0\%} & 43.4\%
& 6.44 & 3.80\pct{+7.8\%} & 10.23\pct{+190.5\%} & 65.6\%
& \underline{0} & 4.24\pct{+20.4\%} & \underline{4.24}\pct{+20.4\%} & \underline{0\%}
& \textbf{0} & \textbf{3.66}\pct{+3.9\%} & \textbf{3.66}\pct{+3.9\%} & \textbf{0\%} \\
\midrule
Electricity
& 0 & 3.51 & 3.51 & 0\%
& 1.49 & \underline{3.79}\pct{+8.0\%} & 5.28\pct{+50.4\%} & 31.2\%
& 4.62 & 3.79\pct{+8.2\%} & 8.41\pct{+139.7\%} & 57.9\%
& \underline{0} & 3.89\pct{+11.0\%} & \underline{3.89}\pct{+11.0\%} & \underline{0\%}
& \textbf{0} & \textbf{3.64}\pct{+3.8\%} & \textbf{3.64}\pct{+3.8\%} & \textbf{0\%} \\
\midrule
Exchange
& 0 & 3.53 & 3.53 & 0\%
& 0.51 & \underline{3.79}\pct{+7.5\%} & 4.30\pct{+21.9\%} & 13.3\%
& 3.07 & 3.79\pct{+7.6\%} & 6.87\pct{+94.7\%} & 47.7\%
& \underline{0} & 4.29\pct{+21.7\%} & \underline{4.29}\pct{+21.7\%} & \underline{0\%}
& \textbf{0} & \textbf{3.66}\pct{+3.8\%} & \textbf{3.66}\pct{+3.8\%} & \textbf{0\%} \\
\midrule
Traffic
& 0 & 3.51 & 3.51 & 0\%
& 1.04 & \underline{3.79}\pct{+8.0\%} & 4.82\pct{+37.5\%} & 23.8\%
& 3.90 & 3.79\pct{+8.1\%} & 7.69\pct{+119.2\%} & 53.7\%
& \underline{0} & 3.88\pct{+10.7\%} & \underline{3.88}\pct{+10.7\%} & \underline{0\%}
& \textbf{0} & \textbf{3.64}\pct{+3.8\%} & \textbf{3.64}\pct{+3.8\%} & \textbf{0\%} \\
\midrule
Weather
& 0 & 3.52 & 3.52 & 0\%
& 2.71 & \underline{3.79}\pct{+7.7\%} & 6.49\pct{+84.6\%} & 44.7\%
& 6.67 & 3.79\pct{+7.8\%} & 10.47\pct{+197.6\%} & 66.4\%
& \underline{0} & 3.96\pct{+12.6\%} & \underline{3.96}\pct{+12.6\%} & \underline{0\%}
& \textbf{0} & \textbf{3.65}\pct{+3.9\%} & \textbf{3.65}\pct{+3.9\%} & \textbf{0\%} \\
\bottomrule
\end{tabular}}
\label{tab:chronosbolt_time_latency_compact}

\end{center}
\end{table*}

%% file: tables/different-size-of-Chronos-simple.tex
\begin{table*}[!htbp]
\begin{center}
\caption{Scaling study of TS-Memory across ChronosBolt model sizes. We use the same protocol as in Table~\ref{tab:ts_plugmem_compact}. Full results are in 
Table~\ref{tab:chronosbolt_ts_plugmem_full} of Appendix~\ref{appx:complete_results}.}

\begin{small}
\setlength\tabcolsep{3pt}
\renewcommand{\arraystretch}{0.85} 
\scalebox{0.9}{
\begin{tabular}{c|cc|cc|cc|cc|cc|cc|cc|cc}
\toprule
\multirow{1}{*}{Model} &
\multicolumn{4}{c|}{ChronosBolt Base (205M)} &
\multicolumn{4}{c|}{ChronosBolt Small (48M)} &
\multicolumn{4}{c|}{ChronosBolt Mini (21M)} &
\multicolumn{4}{c}{ChronosBolt Tiny (9M)} \\
\cmidrule{1-17}
\multirow{1}{*}{Dataset} &
\multicolumn{2}{c|}{Origin} & \multicolumn{2}{c|}{TS-Memory} &
\multicolumn{2}{c|}{Origin} & \multicolumn{2}{c|}{TS-Memory} &
\multicolumn{2}{c|}{Origin} & \multicolumn{2}{c|}{TS-Memory} &
\multicolumn{2}{c|}{Origin} & \multicolumn{2}{c}{TS-Memory} \\
\midrule
Metric &
MSE & MAE & MSE & MAE &
MSE & MAE & MSE & MAE &
MSE & MAE & MSE & MAE &
MSE & MAE & MSE & MAE \\
\midrule
ETTh1 & \underline{0.448} & \underline{0.419} & \textbf{0.421}\pct{-6.0\%} & \textbf{0.414}\pct{-1.2\%} & \underline{0.463} & \underline{0.422} & \textbf{0.437}\pct{-5.5\%} & \textbf{0.417}\pct{-1.2\%} & \underline{0.445} & \underline{0.421} & \textbf{0.424}\pct{-4.8\%} & \textbf{0.417}\pct{-1.0\%} & \underline{0.447} & \underline{0.422} & \textbf{0.427}\pct{-4.6\%} & \textbf{0.418}\pct{-1.0\%} \\
\midrule
ETTh2 & \underline{0.367} & \underline{0.380} & \textbf{0.354}\pct{-3.4\%} & \textbf{0.375}\pct{-1.1\%} & \underline{0.362} & \underline{0.382} & \textbf{0.357}\pct{-1.5\%} & \textbf{0.379}\pct{-0.7\%} & \underline{0.370} & \underline{0.384} & \textbf{0.358}\pct{-3.2\%} & \textbf{0.380}\pct{-1.0\%} & \underline{0.367} & \underline{0.385} & \textbf{0.361}\pct{-1.7\%} & \textbf{0.383}\pct{-0.7\%} \\
\midrule
ETTm1 & \underline{0.421} & \underline{0.383} & \textbf{0.381}\pct{-9.4\%} & \textbf{0.371}\pct{-3.1\%} & \underline{0.420} & \underline{0.385} & \textbf{0.377}\pct{-10.3\%} & \textbf{0.373}\pct{-3.1\%} & \underline{0.417} & \underline{0.385} & \textbf{0.374}\pct{-10.3\%} & \textbf{0.372}\pct{-3.2\%} & \underline{0.407} & \underline{0.385} & \textbf{0.373}\pct{-8.4\%} & \textbf{0.375}\pct{-2.7\%} \\
\midrule
ETTm2 & \underline{0.291} & \underline{0.317} & \textbf{0.267}\pct{-8.0\%} & \textbf{0.311}\pct{-2.1\%} & \underline{0.284} & \underline{0.317} & \textbf{0.266}\pct{-6.4\%} & \textbf{0.311}\pct{-1.7\%} & \underline{0.290} & \underline{0.319} & \textbf{0.266}\pct{-8.2\%} & \textbf{0.312}\pct{-2.2\%} & \underline{0.284} & \underline{0.317} & \textbf{0.264}\pct{-7.2\%} & \textbf{0.312}\pct{-1.8\%} \\
\midrule
Electricity & \underline{0.159} & \underline{0.244} & \textbf{0.155}\pct{-2.5\%} & \textbf{0.241}\pct{-1.0\%} & \underline{0.163} & \underline{0.249} & \textbf{0.159}\pct{-2.5\%} & \textbf{0.246}\pct{-1.1\%} & \underline{0.167} & \underline{0.253} & \textbf{0.163}\pct{-2.5\%} & \textbf{0.251}\pct{-0.8\%} & \underline{0.172} & \underline{0.259} & \textbf{0.166}\pct{-3.1\%} & \textbf{0.256}\pct{-1.2\%} \\
\midrule
Traffic & \underline{0.435} & \underline{0.263} & \textbf{0.425}\pct{-2.3\%} & \textbf{0.258}\pct{-1.6\%} & \underline{0.412} & \underline{0.264} & \textbf{0.408}\pct{-1.1\%} & \textbf{0.260}\pct{-1.4\%} & \underline{0.417} & \underline{0.269} & \textbf{0.411}\pct{-1.2\%} & \textbf{0.265}\pct{-1.3\%} & \underline{0.419} & \underline{0.276} & \textbf{0.414}\pct{-1.2\%} & \textbf{0.271}\pct{-2.1\%} \\
\midrule
Weather & \underline{0.263} & \underline{0.276} & \textbf{0.239}\pct{-9.1\%} & \textbf{0.267}\pct{-3.5\%} & \underline{0.256} & \underline{0.270} & \textbf{0.239}\pct{-6.5\%} & \textbf{0.266}\pct{-1.8\%} & \underline{0.268} & \underline{0.283} & \textbf{0.240}\pct{-10.6\%} & \textbf{0.273}\pct{-3.4\%} & \underline{0.271} & \underline{0.286} & \textbf{0.239}\pct{-11.7\%} & \textbf{0.275}\pct{-3.8\%} \\
\bottomrule
\end{tabular}}
\end{small}
\label{tab:chronosbolt_ts_memory_compact}
\vspace{-0.5em}
\end{center}
\end{table*}

%% file: tables/plugmem_merged_table-simple.tex
\begin{table*}[!htbp]
\begin{center}
\caption{Cross-model transfer results of TS-Memory across retrieval teachers and frozen TSFM backbones. We use the same protocol as in Table~\ref{tab:ts_plugmem_compact}. Full results are in Table~\ref{tab:plugmem_merged} of Appendix~\ref{appx:complete_results}.
}

\begin{small}
\renewcommand{\arraystretch}{0.80} 
\setlength\tabcolsep{2pt}
\resizebox{\textwidth}{!}{%
\begin{tabular}{c|cc|cc|cc|cc|cc|cc||cc|cc|cc|cc|cc|cc}
\toprule
\multirow{2}{*}{Model} & \multicolumn{12}{c||}{Chronos2-PlugMem} & \multicolumn{12}{c}{Sundial-PlugMem} \\
\cmidrule{2-25}
 &
\multicolumn{4}{c|}{ChronosBolt (base)} &
\multicolumn{4}{c|}{Sundial (base)} &
\multicolumn{4}{c||}{TimesFM (base)} &
\multicolumn{4}{c|}{ChronosBolt (base)} &
\multicolumn{4}{c|}{Chronos2 (base)} &
\multicolumn{4}{c}{TimesFM (base)} \\
\midrule
Dataset &
\multicolumn{2}{c|}{Origin} & \multicolumn{2}{c|}{TS-Memory} &
\multicolumn{2}{c|}{Origin} & \multicolumn{2}{c|}{TS-Memory} &
\multicolumn{2}{c|}{Origin} & \multicolumn{2}{c||}{TS-Memory} &
\multicolumn{2}{c|}{Origin} & \multicolumn{2}{c|}{TS-Memory} &
\multicolumn{2}{c|}{Origin} & \multicolumn{2}{c|}{TS-Memory} &
\multicolumn{2}{c|}{Origin} & \multicolumn{2}{c}{TS-Memory} \\
\midrule
Metric &
MSE & MAE & MSE & MAE &
MSE & MAE & MSE & MAE &
MSE & MAE & MSE & MAE &
MSE & MAE & MSE & MAE &
MSE & MAE & MSE & MAE &
MSE & MAE & MSE & MAE \\
\midrule

ETTh1 & \underline{0.448} & \underline{0.419} & \textbf{0.420} & \textbf{0.416} & \underline{0.400} & \underline{0.412} & \textbf{0.395} & \textbf{0.410} & \underline{0.479} & \underline{0.442} & \textbf{0.436} & \textbf{0.430} & \underline{0.448} & \underline{0.419} & \textbf{0.409} & \textbf{0.416} & \underline{0.442} & \textbf{0.412} & \textbf{0.409} & \underline{0.412} & \underline{0.479} & \underline{0.442} & \textbf{0.424} & \textbf{0.427} \\
\midrule
ETTh2 & \underline{0.366} & \textbf{0.380} & \textbf{0.342} & \underline{0.383} & \underline{0.344} & \underline{0.380} & \textbf{0.338} & \textbf{0.379} & \underline{0.402} & \underline{0.409} & \textbf{0.358} & \textbf{0.402} & \underline{0.366} & \textbf{0.380} & \textbf{0.339} & \underline{0.382} & \underline{0.376} & \textbf{0.383} & \textbf{0.347} & \underline{0.384} & \underline{0.402} & \underline{0.409} & \textbf{0.356} & \textbf{0.399} \\
\midrule
ETTm1 & \underline{0.421} & \underline{0.383} & \textbf{0.375} & \textbf{0.375} & \underline{0.369} & \underline{0.369} & \textbf{0.356} & \textbf{0.366} & \underline{0.429} & \underline{0.416} & \textbf{0.375} & \textbf{0.390} & \underline{0.421} & \underline{0.383} & \textbf{0.367} & \textbf{0.375} & \underline{0.433} & \underline{0.381} & \textbf{0.370} & \textbf{0.378} & \underline{0.429} & \underline{0.416} & \textbf{0.372} & \textbf{0.390} \\
\midrule
ETTm2 & \underline{0.290} & \underline{0.317} & \textbf{0.262} & \textbf{0.314} & \underline{0.276} & \underline{0.317} & \textbf{0.262} & \textbf{0.314} & \underline{0.332} & \underline{0.341} & \textbf{0.267} & \textbf{0.322} & \underline{0.290} & \underline{0.317} & \textbf{0.265} & \textbf{0.314} & \underline{0.295} & \underline{0.315} & \textbf{0.264} & \textbf{0.313} & \underline{0.332} & \underline{0.341} & \textbf{0.273} & \textbf{0.325} \\
\midrule
Electricity & \underline{0.159} & \underline{0.244} & \textbf{0.154} & \textbf{0.242} & \underline{0.148} & \underline{0.242} & \textbf{0.144} & \textbf{0.238} & \underline{0.154} & \underline{0.244} & \textbf{0.151} & \textbf{0.242} & \underline{0.159} & \underline{0.244} & \textbf{0.154} & \textbf{0.243} & \underline{0.163} & \underline{0.244} & \textbf{0.157} & \textbf{0.242} & \underline{0.154} & \underline{0.244} & \textbf{0.151} & \textbf{0.243} \\
\midrule
Exchange & \underline{0.371} & \underline{0.412} & \textbf{0.355} & \textbf{0.405} & \underline{0.553} & \underline{0.494} & \textbf{0.408} & \textbf{0.444} & \underline{0.433} & \underline{0.446} & \textbf{0.383} & \textbf{0.427} & \underline{0.371} & \underline{0.412} & \textbf{0.365} & \textbf{0.407} & \underline{0.399} & \underline{0.421} & \textbf{0.392} & \textbf{0.417} & \underline{0.433} & \underline{0.446} & \textbf{0.420} & \textbf{0.438} \\
\midrule
Traffic & \underline{0.435} & \underline{0.263} & \textbf{0.417} & \textbf{0.260} & \underline{0.461} & \underline{0.286} & \textbf{0.427} & \textbf{0.275} & \underline{0.370} & \underline{0.244} & \textbf{0.366} & \textbf{0.243} & \underline{0.435} & \underline{0.263} & \textbf{0.418} & \textbf{0.261} & \underline{0.394} & \textbf{0.237} & \textbf{0.388} & \underline{0.240} & \underline{0.370} & \underline{0.244} & \textbf{0.367} & \textbf{0.243} \\
\midrule
Weather & \underline{0.263} & \underline{0.276} & \textbf{0.235} & \textbf{0.269} & \underline{0.244} & \underline{0.269} & \textbf{0.235} & \textbf{0.269} & \underline{0.222} & \underline{0.237} & \textbf{0.198} & \textbf{0.232} & \underline{0.263} & \underline{0.276} & \textbf{0.235} & \textbf{0.273} & \underline{0.274} & \textbf{0.267} & \textbf{0.233} & \underline{0.268} & \underline{0.222} & \underline{0.237} & \textbf{0.198} & \textbf{0.233} \\

\bottomrule
\end{tabular}}
\end{small}
\label{tab:plugmem_merged_compact}
\end{center}

\end{table*}

%% file: tables/Train-test-domain-configurations.tex
\newcommand{\TabScale}{0.70}   
\newcommand{\TabStretch}{1} 

\begin{table}[t]
    \centering
    \small
    \setlength{\tabcolsep}{4.5pt}
    \renewcommand{\arraystretch}{\TabStretch} 
    \caption{Train$\rightarrow$Test dataset pairs under each setting. $\emptyset$: Zero Shot; ETT$_{\text{pool}}$: pooled four ETT datasets.}

    \label{tab:setting_test_relation}
    \scalebox{\TabScale}{
    \begin{tabular}{lcccc}
        \toprule
        \textbf{Setting} & \multicolumn{4}{c}{\textbf{Train Dataset} $\rightarrow$ \textbf{Test Dataset}} \\
        \midrule
        Origin 
        & $\emptyset \rightarrow$ ETTh1 
        & $\emptyset \rightarrow$ ETTh2 
        & $\emptyset \rightarrow$ ETTm1 
        & $\emptyset \rightarrow$ ETTm2 \\
        Cross-Dom 
        & Weather $\rightarrow$ ETTh1 
        & Weather $\rightarrow$ ETTh2 
        & Weather $\rightarrow$ ETTm1 
        & Weather $\rightarrow$ ETTm2 \\
        Dom-Shift 
        & ETTh2 $\rightarrow$ ETTh1 
        & ETTh1 $\rightarrow$ ETTh2 
        & ETTm2 $\rightarrow$ ETTm1 
        & ETTm1 $\rightarrow$ ETTm2 \\
        Multi-Dom 
        & ETT$_{\text{pool}} \rightarrow$ ETTh1 
        & ETT$_{\text{pool}} \rightarrow$ ETTh2 
        & ETT$_{\text{pool}} \rightarrow$ ETTm1 
        & ETT$_{\text{pool}} \rightarrow$ ETTm2 \\
        In-Dom 
        & ETTh1 $\rightarrow$ ETTh1 
        & ETTh2 $\rightarrow$ ETTh2 
        & ETTm1 $\rightarrow$ ETTm1 
        & ETTm2 $\rightarrow$ ETTm2 \\
        \bottomrule
    \end{tabular}%
    }
    \vspace{-1em}
\end{table}

%% file: contents/5_conclusion.tex
\section{Conclusion and Future Work}

We introduce \textbf{Parametric Memory Distillation}, a retrieval-to-memory paradigm that compiles retrieval-conditioned predictive distributions into a lightweight, plug-and-play module for adapting frozen Time Series Foundation Models. Realized as \textbf{TS-Memory}, our approach distills confidence-aware quantile corrections from an offline, leakage-safe $k$NN teacher and enables retrieval-free inference through simple fusion with the frozen backbone. Experiments across multiple TSFM backbones and long-horizon benchmarks demonstrate consistent improvements in both point and probabilistic forecasting with negligible inference overhead. Future directions include continual memory updates under evolving domains and richer teacher constructions with improved reliability estimation.

\section*{Limitations}

TS-Memory focuses on long-horizon probabilistic forecasting; extending it to imputation, anomaly detection, and decision-focused forecasting remains future work. 
Its shift alignment mainly handles additive channel-wise offsets, so temporal warping, frequency shifts, or abrupt regime changes may reduce reliability. 
When the retrieval corpus poorly matches the deployment domain, teacher targets can be noisy; the advantage gate, anchoring loss, and backbone fusion mitigate but cannot eliminate this risk. 
We use a validation-tuned fusion weight for each dataset--backbone pair at test time, leaving input-adaptive fusion for future study.

\section*{Acknowledgement}

We thank the reviewers for their valuable comments and efforts to improve this manuscript. This work is mainly supported by the Guangdong Basic and Applied Basic Research Foundation (No. 2025A1515011994). This work is also supported by the National Natural Science Foundation of China (No. 62402414), Guangdong Provincial Project 2025D03J0014, Guangzhou Municipal Science and Technology Project (No. 2023A03J0011), the Guangzhou Industrial Information and Intelligent Key Laboratory Project (No. 2024A03J0628), and Guangdong Provincial Key Lab of Integrated Communication, Sensing and Computation for Ubiquitous Internet of Things (No. 2023B1212010007). Raymond Chi-Wing Wong was supported by the fund PRP/004/25FX.

%% file: contents/Appendix.tex
\section*{Appendix}

\input{appendix/A_Method_Supplement}

\input{appendix/B_Experimental_Details}

\input{appendix/C_Experiments_Full_Results}

\input{appendix/D_Future_Work}

%% file: appendix/A_Method_Supplement.tex
\section{Leakage-Safe Retrieval Teacher Construction}
\label{appx:Offline_Teacher_Construction_Algorithm}

\input{tables/Offline-Teacher-Construction}

\section{Theoretical Analysis}
\label{sec:theory}

TS-Memory aims to capture the \emph{adaptivity} of retrieval-based forecasting without paying retrieval-time
latency at deployment. This appendix provides intuition from three complementary views:
\begin{enumerate}
    \item \textbf{Privileged distributional supervision.} Offline retrieval induces a discrete,
    context-conditioned predictive distribution. The teacher quantiles $\widetilde{\mathbf{Q}}_t$
    are weighted-quantile solutions of a pinball objective, linking our targets to local
    non-parametric conditional quantile estimation (Section~\ref{subsec:teacher}).
    \item \textbf{Shift-aligned retrieval.} The alignment in Eq.~(\ref{eq:tsm-shift})--(\ref{eq:tsm-align})
    and reranking explicitly handle additive level shifts, so neighbors are compared mainly by
    \emph{shape} rather than absolute scale (Section~\ref{subsec:shift}).
    \item \textbf{Conservative distillation and deployment.} Reliability-aware gating reduces negative
    transfer by down-weighting (or disabling) distillation when retrieval is unreliable, and
    deployment-time quantile fusion is a convex combination with an immediate pinball-risk guarantee
    (Section~\ref{subsec:reliability}--\ref{subsec:deployment}).
\end{enumerate}

\subsection{Offline Retrieval Supervision}
\label{subsec:teacher}

Given a query context $\mathbf{X}_t$, let $\{\mathbf{Y}^{(k)}_{\text{align}}\}_{k=1}^{K}$ be the top-$K$
aligned future windows retrieved from the \emph{training-only} index, with weights
$\{w_k\}_{k=1}^{K}$ such that $\sum_{k=1}^{K} w_k = 1$ (Eq.~\ref{eq:tsm-ret-weights}).
This defines a teacher available only during offline training; the memory module learns to reproduce its
distributional information without test-time search.
For each horizon--variable index $u\in\mathcal{U}$, define samples
$v_k \triangleq Y^{(k)}_{\text{align},u}$ and the weighted empirical measure
\begin{equation}
\widehat{\mathcal{P}}_{t,u}
\triangleq
\sum_{k=1}^{K} w_k \,\delta_{v_k},
\end{equation}
which is a discrete predictive distribution supported on retrieved futures. It naturally captures uncertainty
(and even multi-modality) from diverse neighbors, providing richer supervision than a single point label.

\noindent\textbf{Weighted quantiles as pinball minimizers.}
For any quantile level $q\in(0,1)$, consider the weighted pinball risk
\begin{equation}
R_{t,u}(z;q)
\triangleq
\mathbb{E}_{V\sim \widehat{\mathcal{P}}_{t,u}}
\big[\rho_{q}(z,V)\big]
=
\sum_{k=1}^{K} w_k \,\rho_{q}(z,v_k),
\end{equation}
where $\rho_q(z,v)=(v-z)\big(q-\mathbb{I}_{v<z}\big)$.
By a standard subgradient argument, any minimizer $z^\star$ satisfies
\begin{equation}
\sum_{k: v_k < z^\star} w_k \le q \le \sum_{k: v_k \le z^\star} w_k.
\label{eq:weighted-quantile-condition}
\end{equation}
If $\widehat{\mathcal{P}}_{t,u}$ has atoms, the minimizer can be non-unique; Eq.~(\ref{eq:tsm-teacher-quantile})
chooses the \emph{left} (lower) weighted quantile satisfying Eq.~(\ref{eq:weighted-quantile-condition}).
Thus, distilling $\widetilde{\mathbf{Q}}_t$ into $g_{\phi}(\mathbf{X}_t)$ can be viewed as \emph{amortizing} a
local non-parametric conditional quantile estimator $\{(v_k,w_k)\}_{k=1}^{K}$ into a constant-time module.

\noindent\textbf{Kernel view and distributional metrics.}
The teacher can also be interpreted as a kernel-smoothed local estimate of the conditional predictive
distribution around $\mathbf{X}_t$ in the frozen embedding space, since
$w_k \propto \exp(-\psi(d_k)/\tau_{\text{ret}})$.
Moreover, many distributional scores can be expressed (exactly or approximately) as integrals of pinball
losses over quantile levels (e.g., CRPS for univariate targets). Hence, improving quantile alignment across
$\{q_j\}$ tends to improve distributional evaluation.

\subsection{Shift Alignment and Reranking}
\label{subsec:shift}

A common non-stationarity is a channel-wise level offset. We model each window as
\[
\mathbf{X}_t = \mathbf{Z}_t + \mathbf{b}_t,
\qquad
\mathbf{Y}_t = \mathbf{W}_t + \mathbf{b}_t,
\]
where $\mathbf{Z}_t,\mathbf{W}_t$ are shift-free patterns and $\mathbf{b}_t\in\mathbb{R}^{C}$ is a per-window
offset (constant within the context and forecast horizon).

For a retrieved candidate $i$ with offset $\mathbf{b}_i$, the trailing-mean shift in Eq.~(\ref{eq:tsm-shift})
estimates $\mathbf{b}_t-\mathbf{b}_i$:
\[
\boldsymbol{s}_i
=
\mathrm{mean}\!\left(\mathbf{X}_{t,L-m+1:L}\right)
-
\mathrm{mean}\!\left(\mathbf{X}^{(i)}_{L-m+1:L}\right)
\approx
\mathbf{b}_t-\mathbf{b}_i.
\]
After applying Eq.~(\ref{eq:tsm-align}), we obtain approximately
\begin{equation}
\mathbf{X}^{(i)}_{\text{align}} \approx \mathbf{Z}^{(i)} + \mathbf{b}_t,
\qquad
\mathbf{Y}^{(i)}_{\text{align}} \approx \mathbf{W}^{(i)} + \mathbf{b}_t,
\end{equation}
so retrieved futures are expressed in the same offset frame as the query.

\noindent\textbf{Why reranking matters.}
TS-Memory reranks by $\text{score}_i=\|\mathbf{X}_t-\mathbf{X}^{(i)}_{\text{align}}\|_1$.
Under the additive-shift model,
\[
\mathbf{X}_t-\mathbf{X}^{(i)}_{\text{align}}
\approx
\mathbf{Z}_t-\mathbf{Z}^{(i)},
\]
so reranking compares \emph{shift-free shapes} rather than absolute levels, reducing scale-mismatch artifacts.
This complements Instance Normalization in Eq.~(\ref{eq:tsm-inorm}): normalization encourages shift-invariant
learning, while alignment (and inverse normalization) restores outputs to the correct absolute scale.

\subsection{Reliability-Aware Distillation}
\label{subsec:reliability}

Retrieval-based distillation can hurt when neighbors are noisy or mismatched. TS-Memory reduces this risk
with a two-stage reliability mechanism: (i) a confidence proxy from retrieval separation, and (ii) an
advantage test that activates distillation only when the teacher is better than the frozen backbone.

\noindent\textbf{Confidence as retrieval separation.}
We use $\mathrm{Conf}_t=\max_k w_k$ (Eq.~\ref{eq:tsm-conf}). Let $d_{(1)}\le d_{(2)}\le\cdots\le d_{(K)}$ be
sorted distances and $\Delta=d_{(2)}-d_{(1)}$ the top-1 margin. Since the weights are a softmax over distances,
\begin{equation}
\mathrm{Conf}_t = w_{(1)}
\ge \frac{1}{1+(K-1)\exp(-\Delta/\tau_{\text{ret}})}.
\label{eq:conf-margin}
\end{equation}
Larger $\mathrm{Conf}_t$ indicates a clearer nearest-neighbor match.

\noindent\textbf{Distill only when beneficial.}
Confidence alone can be misleading, so TS-Memory further gates distillation using an advantage test
(Eq.~\ref{equ:err}):
\begin{equation}
\chi_t
=
\mathbb{I}\!\left(
\mathrm{err}^{T}_t + \epsilon_{\text{gate}}
<
\mathrm{err}^{\text{base}}_t
\right),
\qquad
\omega_t
=
\chi_t \cdot \mathrm{Conf}_t^{\gamma}.
\end{equation}
The alignment loss (Eq.~\ref{eq:tsm-loss-align}) is weighted by $\omega_t$; when $\chi_t=0$, distillation
contributes no gradient, avoiding systematic negative transfer.

\noindent\textbf{Residual distillation and stabilization.}
Besides matching teacher quantiles (Eq.~\ref{eq:tsm-dq}), TS-Memory distills an incremental median correction
over the frozen backbone (Eq.~\ref{eq:tsm-delta}--\ref{eq:tsm-ddelta}):
\[
\Delta^{\text{mem}}_{t,u}
=
\widehat{Q}^{\text{mem}}_{t,j^\star,u}
-
\widehat{Q}^{\text{base}}_{t,j^\star,u},
\qquad
\Delta^{T}_{t,u}
=
\widetilde{Q}_{t,j^\star,u}
-
\widehat{Q}^{\text{base}}_{t,j^\star,u}.
\]
This focuses learning on the teacher's \emph{improvement} over a strong frozen prior. When retrieval is
uncertain, the anchoring loss (Eq.~\ref{eq:tsm-loss-anchor}) pulls the memory median toward the backbone
median with weight $(1-\omega_t)$, and the quantile crossing penalty (Eq.~\ref{eq:tsm-loss-cross}) encourages
monotonicity across quantile levels.

\subsection{Deployment via Convex Quantile Fusion}
\label{subsec:deployment}

At inference time, TS-Memory performs no retrieval and fuses backbone and memory forecasts by quantile-wise
interpolation (Eq.~\ref{eq:tsm-fuse}):
\begin{equation}
\widehat{\mathbf{Q}}^{\text{final}}_t
=
(1-\alpha)\widehat{\mathbf{Q}}^{\text{base}}_t
+
\alpha \widehat{\mathbf{Q}}^{\text{mem}}_t,
\qquad
\alpha\in[0,1].
\end{equation}
We tune a single $\alpha$ on validation to preserve quantile coherence and keep tuning 1D.

Because $\rho_q(z,y)$ is convex in $z$, for any fixed $(t,j,u)$ and target $y$,
\begin{equation}
\begin{aligned}
\rho_{q_j}\!\Big(
(1-\alpha)\widehat{Q}^{\text{base}}_{t,j,u}
+
\alpha \widehat{Q}^{\text{mem}}_{t,j,u},
\, y
\Big)
\le\;&
(1-\alpha)\rho_{q_j}\!\Big(\widehat{Q}^{\text{base}}_{t,j,u},y\Big) \\
&+
\alpha \rho_{q_j}\!\Big(\widehat{Q}^{\text{mem}}_{t,j,u},y\Big).
\end{aligned}
\end{equation}
Averaging over horizons, variables, and quantiles yields that the fused empirical pinball risk is
upper-bounded by the same convex combination of the individual risks. Therefore, the best $\alpha\in[0,1]$
(on any evaluation set) is never worse than choosing either endpoint predictor. In practice, since the
objective is convex (piecewise-linear) in $\alpha$, a small grid search on validation pinball/CRPS is enough.
Since the risk is convex in $\alpha$, any local optimum is global; increasing grid resolution only refines the minimizer.
If both quantile sets are monotone in $q_j$, their convex combination remains monotone, so fusion preserves
distribution validity.

\noindent\textbf{Extension to point forecasting.}
When point forecasts are needed, we extract a point estimate from quantiles. Let $q_{j^\star}$ be closest to
$0.5$. If $0.5\in\mathcal{Q}$, use the median; otherwise interpolate between $q_{j_a}<0.5<q_{j_b}$:
\begin{equation}
\widehat{p}_{t,u}
=
\widehat{Q}_{t,j_a,u}
+
\frac{0.5-q_{j_a}}{q_{j_b}-q_{j_a}}
\Big(
\widehat{Q}_{t,j_b,u}
-
\widehat{Q}_{t,j_a,u}
\Big).
\end{equation}
Point forecast fusion is analogous:
\begin{equation}
\widehat{p}^{\text{final}}_{t,u}
=
(1-\alpha)\widehat{p}^{\text{base}}_{t,u}
+
\alpha \widehat{p}^{\text{mem}}_{t,u},
\end{equation}
with $\alpha$ tuned on validation MAE/MSE. For point-only backbones, treat the backbone output as
$\widehat{p}^{\text{base}}$.

%% file: tables/Offline-Teacher-Construction.tex
\begin{algorithm}[!htbp]
\caption{Offline Teacher Construction}
\label{alg:teacher}
\begin{algorithmic}[1]
\Require training split $\mathcal{D}_{\text{train}}$, quantile levels $\mathcal{Q}$, neighbors $K$, trailing length $m$,
        frozen encoder $f_{\text{enc}}$, distance transform $\psi(\cdot)$, temperature $\tau_{\text{ret}}$
\Ensure teacher dataset $\mathcal{D}_{\text{teach}}$

\State $\mathcal{K} \gets \mathcal{D}_{\text{train}}$ \Comment{leakage-safe knowledge base}
\State Pre-compute $\mathbf{e}_i \gets f_{\text{enc}}(\mathbf{X}^{(i)})$ for all $(\mathbf{X}^{(i)},\mathbf{Y}^{(i)})\in\mathcal{K}$; build an NN index on $\{\mathbf{e}_i\}$
\State $\mathcal{D}_{\text{teach}} \gets \emptyset$

\For{each supervised window $(\mathbf{X}_t,\mathbf{Y}_t)$ in $\mathcal{K}$}
    \State Query NN index with $\mathbf{e}_t \gets f_{\text{enc}}(\mathbf{X}_t)$ to get candidates $\mathcal{N}_t$ and distances $\{d_k\}$; drop self-matches
    \State Align candidates by shifting the last-$m$ context mean; rerank by aligned-context $\ell_1$ and keep top-$K$
    \State $w_k \gets \mathrm{Softmax}\!\big(-\psi(d_k)/\tau_{\text{ret}}\big)$; $\mathrm{Conf}_t \gets \max_k w_k$
    \State $\widetilde{\mathbf{Q}}_t \gets$ weighted empirical quantiles of aligned futures under $\{w_k\}$ at levels $\mathcal{Q}$
    \State $\mathcal{D}_{\text{teach}} \gets \mathcal{D}_{\text{teach}} \cup \{(\mathbf{X}_t,\mathbf{Y}_t,\widetilde{\mathbf{Q}}_t,\mathrm{Conf}_t)\}$
\EndFor

\State \Return $\mathcal{D}_{\text{teach}}$
\end{algorithmic}
\end{algorithm}

%% file: appendix/B_Experimental_Details.tex
\section{Experimental Details}
\label{appx:experiment_details}

\subsection{Datasets and splits}

We evaluate our method on eight widely used benchmark datasets that collectively cover a broad range of real-world time series forecasting scenarios (Table~\ref{tab:dataset}). The benchmarks span diverse application domains, including electricity transformer temperature monitoring (ETTm1, ETTm2, ETTh1, ETTh2), power consumption analysis (Electricity), transportation systems (Traffic), meteorological forecasting (Weather), and foreign exchange markets (Exchange-rate). All datasets contain multivariate time series with different dimensionalities and sequence lengths, and are split into training/validation/testing sets following standard protocols. They also exhibit heterogeneous temporal characteristics: the sampling frequency ranges from 15-minute intervals to daily observations, and the underlying series present distinct periodic patterns that reflect real-world dynamics.

\input{tables/dataset-details}

\noindent\textbf{Dataset descriptions.}
\begin{itemize}[leftmargin=*]
\item \textbf{ETT}: Four datasets (ETTh1, ETTh2, ETTm1, ETTm2) containing two years of electricity transformer temperature data collected from two counties in China. ETTh1/ETTh2 provide hourly measurements, while ETTm1/ETTm2 have 15-minute resolution. Each dataset contains seven variables: six power load features and one target oil temperature variable.
  \item \textbf{Traffic}: Hourly road occupancy rates from 862 sensors deployed on the freeway system in the San Francisco Bay Area. The dataset captures traffic flow patterns and congestion dynamics across multiple road segments.
  \item \textbf{Weather}: Meteorological measurements from 21 weather stations in Germany, recorded every 10 minutes over one year. The dataset includes 21 atmospheric indicators such as air temperature, humidity, atmospheric pressure, and wind conditions.
  \item \textbf{Electricity}: Hourly electricity consumption records from 321 customers (residential and commercial). The dataset exhibits complex daily and seasonal usage patterns driven by heterogeneous user behaviors.
  \item \textbf{Exchange-rate}: A multivariate foreign exchange dataset with daily exchange rates of eight currencies against the U.S.\ dollar. It contains 7,588 time steps and 8 variables (Australia, UK, Canada, Switzerland, China, Japan, New Zealand, and Singapore). This dataset is widely used to benchmark long-horizon multivariate forecasting under non-stationarity and regime shifts.
\end{itemize}

\noindent\textbf{Periodicity encoding.}
The \textit{Periodicity} column in Table~\ref{tab:dataset} reports the dataset-specific periodicity hyperparameter $P$ used in our periodicity encoding. This parameter is chosen to reflect the dominant cyclic structure of each dataset (e.g., daily or weekly cycles) given its sampling frequency. Concretely, ETTm1/ETTm2 (15-minute sampling) use $P=96$ to represent one day ($24\times4$ samples), while ETTh1/ETTh2, Electricity, and Traffic (hourly sampling) use $P=24$ for daily periodicity. Weather (10-minute sampling) uses $P=144$ ($24\times6$ samples) to capture a full day. Exchange-rate (daily sampling) uses $P=7$ to model weekly periodicity. We adopt the following trigonometric formulation:
\begin{equation}
  \text{encoding}(t)=\left[\sin\left(\frac{2\pi t}{P}\right),\ \cos\left(\frac{2\pi t}{P}\right)\right],
\end{equation}
where $t$ is the discrete time index and $P$ is the dataset-specific period. The resulting sinusoidal features are concatenated with the original time series representation to improve the model’s ability to capture both short-term dependencies and long-range periodic patterns.

\subsection{Evaluation metrics.}
For long-term forecasting, we report Mean Squared Error (MSE), Mean Absolute Error (MAE)~\cite{hyndman2006another}, and the Continuous Ranked Probability Score (CRPS)~\cite{gneiting2007strictly}:
\begin{align}
\text{MSE} &= \frac{1}{H}\sum_{h=1}^{H} \left(\mathbf{Y}_{h}-\hat{\mathbf{Y}}_{h}\right)^2, \label{equ:mse}\\
\text{MAE} &= \frac{1}{H}\sum_{h=1}^{H}\left|\mathbf{Y}_{h}-\hat{\mathbf{Y}}_{h}\right|, \label{equ:mae}\\
\text{CRPS}(\mathcal{D},\mathbf{Y}) &= \int_{\mathbb{R}} \left(F_{\mathcal{D}}(x) - \mathbb{I}\{x \ge \mathbf{Y}\}\right)^2 \, dx. \label{equ:crps}
\end{align}

where $H$ is the prediction horizon, and $\mathbf{Y}_{h}$ and $\hat{\mathbf{Y}}_{h}$ denote the ground-truth and predicted values at step $h$, respectively. For CRPS, $F_{\mathcal{D}}$ denotes the cumulative distribution function of the forecast distribution $\mathcal{D}$ and $\mathbb{I}{\cdot}$ is the Heaviside step (indicator) function. For multivariate series, the squared and absolute operations are applied element-wise and averaged across variables. Following standard practice, we approximate CRPS using the mean weighted quantile loss when only a finite set of predictive quantiles is available~\cite{park2022learning}.

\subsection{Frozen Backbones}
\noindent\textbf{Backbone models and frozen setting.}
We evaluate TS-Memory on four representative Time Series Foundation Models with publicly released pretrained checkpoints, including ChronosBolt~\cite{ansari2024chronos}, TimesFM~\cite{das2024decoder}, Chronos2~\cite{ansari2025chronos}, and Sundial~\cite{liu2025sundial}.
These backbones are chosen (i) to ensure reproducibility, (ii) to cover multiple generations of TSFM development from earlier to more recent releases, and (iii) to provide empirically strong zero-shot forecasters such that any improvement remains meaningful.
All selected TSFMs expose a compatible probabilistic forecasting interface that returns a fixed set of predictive quantiles, enabling plug-and-play adaptation under a unified evaluation setup.
In all experiments, the backbone is strictly frozen.
Given a context window $X_t$, a backbone $f_{\theta}$ outputs quantile forecasts
$\widehat{\mathbf{Q}}^{\text{base}}_{t}=f_{\theta}(X_t) \in \mathbb{R}^{Q \times H \times C}$,
where $Q$ is the number of quantile levels, $H$ is the forecast horizon, and $C$ is the number of variables.
No backbone parameters are updated; any performance change comes solely from the adaptation strategy.

\noindent\textbf{Long-horizon inference protocol.}
Following the standard long-horizon forecasting setup, we evaluate multiple horizons and report averaged results across $H \in \{96,192,336,720\}$ with a fixed look-back window of $L=512$ unless specified otherwise.
When a backbone produces forecasts in shorter blocks, we use a rolling strategy: predicted values are appended to the context and forecasting is repeated until the target horizon is reached.
We apply the same rollout protocol to \textit{Origin}, online retrieval baselines, and TS-Memory to ensure a fair comparison in both accuracy and latency.

\noindent\textbf{ChronosBolt.}
ChronosBolt is included primarily because it provides a comprehensive collection of released pretrained checkpoints across multiple model sizes.
This enables controlled scaling studies under a single adaptation protocol and facilitates accuracy--latency analyses without confounding changes in model families.

\noindent\textbf{TimesFM.}
TimesFM serves as an earlier and widely adopted TSFM baseline that anchors our evaluation.
Including TimesFM ensures that conclusions are not restricted to only the latest model families and helps verify whether the proposed adaptation yields consistent gains on an established backbone.

\noindent\textbf{Chronos2.}
Chronos2 represents more recent advances within the Chronos family and is widely regarded as a strong contemporary TSFM.
Evaluating on Chronos2 helps verify that gains persist on competitive backbones rather than arising only on weaker models.

\noindent\textbf{Sundial.}
Sundial is chosen as another recent and high-performing TSFM that complements the Chronos line and increases backbone diversity.
Covering multiple strong TSFMs reduces the risk that results are overly tied to a single architectural design and highlights the robustness of TS-Memory across backbones.

\subsection{Adaptation Baselines}
\noindent\textbf{Baselines overview.}
We compare TS-Memory with (i) \textit{Origin}, which directly applies each frozen TSFM without adaptation;
(ii) online retrieval augmentation baselines that perform test-time $k$NN retrieval of similar context--future exemplars and fuse retrieved evidence into the final forecast, including TS-RAG~\cite{ning2025ts} and a RAFT-style retrieval baseline~\cite{han2025retrieval}; and
(iii) LoRA~\cite{hu2022lora}, a parameter-efficient fine-tuning baseline on ChronosBolt.

\noindent\textbf{Leakage-safe retrieval corpus.}
For all online retrieval baselines, the retrieval corpus is constructed strictly from the training split to prevent information leakage.
Specifically, we build a knowledge base of context--future pairs
$\mathcal{K}=\{(X^{(i)},Y^{(i)})\}_{i=1}^{N_{\text{train}}}$,
where each indexed window is fully contained inside the training segment.
The same leakage-safe restriction applies when constructing the offline retrieval teacher for TS-Memory, ensuring that online baselines and TS-Memory rely on comparable information.

\noindent\textbf{Common probabilistic fusion interface.}
Both online retrieval baselines return retrieved future windows together with similarity/distance scores.
To enable probabilistic evaluation, we treat retrieved futures as weighted samples, compute retrieved quantiles
$\widehat{\mathbf{Q}}^{\text{ret}}_t \in \mathbb{R}^{Q\times H\times C}$ using weighted empirical quantiles, and fuse them with the TSFM quantile forecast via quantile-wise interpolation:
\begin{equation*}
\widehat{\mathbf{Q}}^{\text{fuse}}_t=(1-\beta)\widehat{\mathbf{Q}}^{\text{base}}_t+\beta\widehat{\mathbf{Q}}^{\text{ret}}_t,\quad \beta\in[0,1].
\end{equation*}
The fusion weight $\beta$ is tuned on the validation split and fixed for test evaluation.

\noindent\textbf{RAFT-style correlation retrieval.}
RAFT~\cite{han2025retrieval} was originally proposed as a standalone retrieval-augmented forecaster; we adapt its retrieval mechanism into an online module compatible with frozen TSFMs.
Given a query context, RAFT computes multi-granularity representations by summarizing non-overlapping blocks at different temporal scales and performing offset removal at the final step.
We retrieve top-$k$ neighbors based on mean-centered cosine similarity averaged across granularities.
Retrieval weights are computed via softmax (temperature tuned on validation), and retrieved futures are converted to quantiles using the common probabilistic interface before fusion with TSFM outputs.

\noindent\textbf{TS-RAG-style embedding retrieval.}
TS-RAG~\cite{ning2025ts} performs nearest neighbor retrieval in an embedding space using a frozen embedding model.
Following the public implementation, we pre-compute embeddings for training contexts and build a FAISS index for fast $k$NN search using $\ell_2$ distance.
At inference, we retrieve top-$k$ neighbors for each query context, reconstruct the corresponding context-plus-future sequences, compute retrieved quantiles with the same weighting scheme, and fuse them with TSFM outputs through the common probabilistic interface.
This baseline represents a typical embedding-based retrieval pipeline, contrasting RAFT's correlation retrieval that does not rely on a learned embedding space.

\noindent\textbf{LoRA fine-tuning.}
LoRA~\cite{hu2022lora} was proposed for parameter-efficient fine-tuning of large language models; we apply it to ChronosBolt as an adaptation baseline.
Following standard practice, we inject low-rank updates into attention projection layers while freezing backbone weights.
Each projection is augmented by a rank-$r$ update scaled by $\alpha/r$, with $r$ chosen to match the trainable-parameter budget of PlugMem.
We initialize injected weights to preserve model behavior at step 0 and train only the low-rank parameters.

\noindent\textbf{Hyper-parameter tuning.}
For both retrieval baselines, we tune the number of neighbors and the fusion weight on the validation split.
For RAFT-style retrieval, we additionally tune the softmax temperature, while keeping the multi-granularity decomposition consistent with the reference implementation.
For LoRA, we tune learning rate and training epochs on validation.
All baselines use the same split and validation protocol to ensure fair comparison.

\subsection{TS-Memory Module}
\noindent\textbf{PlugMem architecture.}
TS-Memory augments a frozen TSFM with a lightweight memory module $g_{\phi}$, referred to as PlugMem.
PlugMem is implemented as an encoder--decoder Transformer that consumes only the context window $X_t$ and does not access backbone internals, enabling plug-and-play integration across different TSFMs.
We apply per-window Instance Normalization on $X_t$, partition the normalized sequence into non-overlapping patches of length $p$,
project patches into $d$-dimensional tokens, and encode them with a Transformer encoder.
A Transformer decoder uses $H$ horizon queries (one per forecast step) to attend over the encoded memory and produce horizon-conditioned features.
A quantile head maps each feature to $Q$ quantile values, and Instance Normalization is inverted to restore the original scale.
While trained with probabilistic outputs, PlugMem also supports point forecasting by taking the median quantile at each horizon as the point estimate, so the same module can serve both probabilistic and point-prediction settings.

\noindent\textbf{Retrieval-free inference and $\alpha$ tuning.}
At test time, TS-Memory is strictly retrieval-free:
we compute $\widehat{\mathbf{Q}}^{\text{base}}_t=f_{\theta}(X_t)$ and $\widehat{\mathbf{Q}}^{\text{mem}}_t=g_{\phi}(X_t)$, then fuse them via quantile-wise interpolation
\begin{equation}
    \widehat{\mathbf{Q}}^{\text{final}}_t=(1-\alpha)\widehat{\mathbf{Q}}^{\text{base}}_t+\alpha \widehat{\mathbf{Q}}^{\text{mem}}_t, \quad \alpha \in [0,1].
\end{equation}
When point forecasts are required, we extract the median quantile from $\widehat{\mathbf{Q}}^{\text{base}}_t$ and $\widehat{\mathbf{Q}}^{\text{mem}}_t$ to obtain point predictions and apply the same interpolation rule.
We tune $\alpha$ on the validation split for each dataset--backbone pair and keep it fixed for test-time evaluation.
Inference requires only two forward passes and introduces no retrieval index maintenance or nearest-neighbor search overhead.

\noindent\textbf{Offline teacher construction (privileged supervision).}
Retrieval is used \emph{only} during offline training as privileged supervision.
For each supervised window $(X_t,Y_t)$, we construct an auxiliary teacher dataset
$\mathcal{D}_{\text{teach}}=\{(X_t,Y_t,\widehat{\mathbf{Q}}^{e}_t,\text{Conf}_t)\}$
from a leakage-safe knowledge base built on the training split.
Following the method in Appendix~A, candidates are retrieved in a frozen embedding space and then shift-aligned via a trailing-mean adjustment over the last $m$ steps to mitigate level offsets.
Aligned candidates are re-ranked, and top-$K$ neighbors are aggregated via softmax weights.
Teacher quantile targets $\widehat{\mathbf{Q}}^{e}_t$ are computed as weighted empirical quantiles over aligned retrieved futures, and retrieval confidence is the concentration of retrieval weights ($\text{Conf}_t=\max_k w_k$).
For point forecasting, the same retrieved futures induce a teacher point target via the weighted empirical median, consistent with using the median quantile as the point estimate.

\noindent\textbf{Training objective.}
PlugMem is trained with frozen backbone, using a composite objective that combines: (i) task supervision on ground-truth futures via quantile regression, (ii) confidence-gated distillation that aligns PlugMem to teacher quantiles more strongly when retrieval is confident and beneficial, and (iii) stability regularization that discourages over-correction and prevents quantile crossing.
Concretely, distillation is activated only when the teacher improves over frozen backbone (with a margin) and is weighted by a confidence-scaled coefficient, while regularization anchors PlugMem to conservative behavior when retrieval is unreliable.
This encourages PlugMem to internalize retrieval benefits when helpful, while remaining robust to noisy or uninformative retrieval.

%% file: tables/dataset-details.tex
\begin{table*}[!htbp]
  \caption{Summary of benchmark datasets. Each dataset includes multiple time series (Dim.) with varying sequence lengths, split into training, validation, and testing sets. Data are collected at different frequencies across various domains.}
  \vspace{-0.5em}
  \label{tab:dataset}
  \centering
  \begin{tabular}{l|c|c|c|c|c|c|c}
    \toprule
    Dataset & Dim. & Forecast Horizons & Dataset Size & Frequency & Domain & Forecastability* & Periodicity \\
    \toprule
    ETTm1 & 7 & {\{96, 192, 336, 720\}} & (34465, 11521, 11521) & 15 min & Temperature & 0.46 & 96 \\
    \cmidrule{1-8}
    ETTm2 & 7 & {\{96, 192, 336, 720\}} & (34465, 11521, 11521) & 15 min & Temperature & 0.55 & 96 \\
    \cmidrule{1-8}
    ETTh1 & 7 & {\{96, 192, 336, 720\}} & (8545, 2881, 2881) & 1 hour & Temperature & 0.38 & 24 \\
    \cmidrule{1-8}
    ETTh2 & 7 & {\{96, 192, 336, 720\}} & (8545, 2881, 2881) & 1 hour & Temperature & 0.45 & 24 \\
    \cmidrule{1-8}
    Electricity & 321 & {\{96, 192, 336, 720\}} & (18317, 2633, 5261) & 1 hour & Electricity & 0.77 & 24 \\
    \cmidrule{1-8}
    Exchange-rate & 8 & {\{96, 192, 336, 720\}} & (5216, 761, 1518) & 1 day & Finance & 0.41 & 7 \\
    \cmidrule{1-8}
    Traffic & 862 & {\{96, 192, 336, 720\}} & (12185, 1757, 3509) & 1 hour & Transportation & 0.68 & 24 \\
    \cmidrule{1-8}
    Weather & 21 & {\{96, 192, 336, 720\}} & (36792, 5271, 10540) & 10 min & Weather & 0.75 & 144 \\
    \bottomrule
    \multicolumn{8}{l}{\scriptsize* Forecastability is computed as $1$ minus the entropy of the Fourier decomposition of a time series \cite{goerg2013forecastable}. Larger values indicate higher predictability.} \\
    
  \end{tabular}
    \vspace{-0.5em}
\end{table*}

%% file: appendix/C_Experiments_Full_Results.tex
\section{Complete results}
\label{appx:complete_results}

\subsection{TS-Memory across frozen TSFM backbones}
\label{appx:different-TSFMs-with-TS-Memory}

Table~\ref{tab:ts_plugmem_full} reports the full horizon-wise MSE/MAE results for TS-Memory on four frozen TSFM backbones (ChronosBolt, Chronos2, Sundial, and TimesFM), complementing the horizon-averaged summary in the main table. Across datasets and horizons, PlugMem yields consistent long-horizon gains without updating backbone parameters. In terms of MSE, average reductions are typically in the 5\%--7\% range for ChronosBolt, Chronos2, and Sundial, while TimesFM often sees larger reductions closer to 7\%--9\%. The complete per-dataset, per-horizon breakdown confirms that improvements are broad-based rather than driven by a small subset of benchmarks or horizons. The horizon-dependent patterns are not uniform across backbones, which motivates reporting the full table. Chronos2 shows a clearer tendency for gains to grow with the forecasting horizon, consistent with intuition that learned memory corrections can mitigate error accumulation during long iterative rollouts. Sundial and TimesFM, by contrast, often realize their strongest relative gains at short and medium horizons while remaining positive at $H=720$. This pattern suggests part of their long-horizon strength may already be captured by their native decoding behavior, with PlugMem providing complementary refinements.

The dataset-level view also reveals heterogeneity that can be obscured by horizon-averaged summaries. Weather and the minute-level ETT benchmarks repeatedly show larger relative improvements, whereas Traffic is more challenging for some backbones. Finally, MAE improvements are generally smaller than MSE improvements, indicating that TS-Memory primarily suppresses occasional large-error outliers rather than uniformly shifting all predictions.

\subsection{Comparison with adaptation baselines}
\label{appx:rag-TS-memory}

\noindent\textbf{Accuracy and calibration.}
Table~\ref{tab:chronosbolt_base_ablation} provides the full horizon-wise comparison on ChronosBolt against online retrieval baselines, including not only point errors (MSE/MAE) but also CRPS. TS-Memory achieves a stable MSE reduction of roughly 5\%--6\% across horizons, while its CRPS improvements become more pronounced as the horizon increases (from around 4\% at $H=96$ to about 8\% at $H=720$). This horizon-dependent CRPS trend is consistent with distillation being particularly beneficial for probabilistic quality when long-range rollouts amplify uncertainty. RAFT yields smaller and flatter gains (around a 3\% MSE reduction with about a 1\% MAE decrease), suggesting that retrieval helps but does not fully correct long-horizon drift. TS-RAG is less stable in the full breakdown: its average MSE reduction remains below about 2.2\% and MAE can degrade at some horizons, implying that directly injecting retrieved trajectories may introduce bias or noise even when it occasionally helps MSE. Dataset-level details reinforce the same message: the strongest distillation benefits concentrate on regimes with clearer repeating structure (e.g., Weather and minute-level ETT), while on Exchange the longest-horizon point-error gains can be small even when CRPS still improves, indicating better-calibrated predictive distributions even when the mean prediction saturates.

\noindent\textbf{Per-horizon latency decomposition.}
Table~\ref{tab:chronosbolt_time_latency} clarifies why online retrieval can be expensive in practice. For both RAFT and TS-RAG, retrieval introduces a substantial overhead and tends to increase with the prediction horizon, consistent with repeated retrieval during rollout or higher effective context cost at longer horizons. At $H=96$, retrieval already accounts for roughly one third of total latency for RAFT and more than half for TS-RAG, corresponding to total slowdowns of about 71\% and 193\% relative to the no-retrieval baseline. As the horizon grows to 720, the forward pass naturally becomes dominant so the retrieval fraction decreases; however, the overall overhead remains large at about 46\% for RAFT and about 126\% for TS-RAG. In contrast, TS-Memory removes retrieval latency entirely and introduces only a small forward-pass overhead (under 6\% across horizons), with the relative overhead becoming less visible for long horizons because decoding cost is unavoidable. Overall, among retrieval-based baselines, distillation is the only approach here that scales to long horizons without making retrieval a dominant runtime bottleneck.

\noindent\textbf{End-to-end evaluation time.}
When timing full evaluation runs rather than individual queries, Table~\ref{tab:chronosbolt_total_time} shows the same trend at wall-clock scale. Even if RAFT slightly reduces forward time in some settings, the added retrieval stage dominates the total runtime, yielding around 57\%--71\% slowdowns at short horizons and still about 40\% extra time at $H=720$. TS-RAG is substantially heavier, producing roughly 118\%--193\% slowdowns depending on horizon while keeping retrieval responsible for about half of the total runtime. TS-Memory remains close to the baseline with only 3.5\%--5.7\% extra total time, and this gap consistently shrinks at larger horizons because the forward computation grows. We note that occasional small speedups on some datasets are largely explained by fixed warm-up overhead and a smaller number of forecasting queries, which makes end-to-end timing less stable. These end-to-end results also highlight a dataset-size effect: on large-scale datasets such as Traffic, the number of forecasting calls is high, so retrieval overhead can accumulate into a prohibitive wall-clock penalty even when the per-query retrieval fraction appears moderate.

\subsection{Transfer generality}
\label{appx:transferability}

\noindent\textbf{Scaling across backbone sizes.}
Shrinking the backbone does not erase the benefit of retrieval distillation, as Table~\ref{tab:chronosbolt_ts_plugmem_full} demonstrates across ChronosBolt sizes. Averaged over horizons and datasets, TS-Memory typically delivers about 4.8\%--5.8\% MSE reduction across sizes, showing that gains are not limited to large backbones. The horizon-wise breakdown reveals an interaction between capacity and long-range forecasting: the mini model exhibits growing relative gains as the horizon increases (from about 4.5\% at $H=96$ to roughly 7.1\% at $H=720$), suggesting that smaller backbones benefit more from learned correction when compounding errors become severe. The tiny model also remains robust, maintaining improvements across all horizons. MAE follows similar but smaller trends, consistent with TS-Memory preferentially removing occasional large deviations rather than uniformly shifting all outputs.

\noindent\textbf{Cross-model transfer across backbones.}
Cross-model transfer in Table~\ref{tab:plugmem_merged} indicates that PlugMem generalizes beyond the specific teacher--target pairing used during distillation. Averaged over datasets and horizons, transfer yields around 7\% MSE reduction on ChronosBolt targets and around 6\% on Sundial targets, while TimesFM targets benefit the most with roughly 9\% MSE reduction. Gains persist across horizons, with TimesFM maintaining strong improvements from $H=96$ through 720, suggesting that the memory captures longer-range error dynamics rather than only short-horizon fixes. Teacher choice is also less critical than one might expect: Chronos2-PlugMem and Sundial-PlugMem produce comparable gains on shared targets, implying that TS-Memory distills a common retrieval signal rather than teacher-specific quirks. The per-dataset view further shows that transfer can be especially valuable under domain difficulty or shift (e.g., large gains on ETTm2 for TimesFM and on Exchange for Sundial), whereas Traffic often sees smaller changes concentrated in MSE rather than MAE.

\noindent\textbf{Transfer under domain shift.}
Table~\ref{tab:domain_split_lora_tsmemory} isolates how the \emph{train–test shift in retrieval supervision} impacts TS-Memory and LoRA: we keep the frozen backbone (ChronosBolt), retrieval hyperparameters, and training recipe fixed, varying only which split/domain provides supervision for the offline retrieval teacher (settings summarized in Table~\ref{tab:setting_test_relation}). A clear pattern emerges: domain alignment is the key driver, and the advantage grows with forecasting horizon. Across all configurations, average MSE reduction widens from $H{=}96$ to $H{=}720$, consistent with longer rollouts relying more heavily on retrieved analogs. Even cross-domain supervision provides measurable transfer, with gains increasing from about $1.4\%$ at $H{=}96$ to roughly $3.7\%$ at $H{=}720$, suggesting that generic temporal motifs can generalize beyond the source domain. Partially aligned domains (distribution-shift or multi-domain) roughly double the benefit, reaching about $4.5\%$--$4.8\%$ MSE reduction at $H{=}720$. In-domain supervision remains strongest, improving MSE by about $5.0\%$ at $H{=}96$ and nearly $7.8\%$ at $H{=}720$. In contrast, LoRA is substantially more sensitive to domain mismatch—it can underperform the frozen backbone under cross-domain training and shows less stable benefits when domains are misaligned. These results demonstrate that retrieval-distilled supervision provides a more robust transfer pathway than parameter-efficient fine-tuning: PlugMem preserves stronger gains under domain shifts by correcting phase/drift errors through external memory without re-fitting backbone parameters.

\noindent\textbf{Scaling PlugMem capacity.}
Following the analysis in Section~\ref{sec:model_analysis}, we evaluate three PlugMem variants (small, base, and large). Table~\ref{tab:mem-size} shows only minor differences across these configurations, with all three landing near a 5.5\% average MSE reduction and about 1.9\%--2.0\% MAE reduction across datasets and horizons. The horizon-level breakdown also indicates that bigger is not uniformly better: the base memory can be slightly stronger at shorter horizons, the large memory can edge out at some medium horizons, and the ordering can flip across horizons, suggesting a non-monotonic interaction between memory capacity and dataset dynamics rather than a simple scaling law. Dataset-level behavior points to the same conclusion: memory size has limited impact on Traffic where improvements remain small, whereas on Weather and minute-level ETT data even the smallest memory delivers substantial gains. Overall, most of the retrieval-distillation benefit is captured with only a few million parameters, making PlugMem tunable for parameter efficiency without sacrificing much accuracy.

%% file: appendix/D_Future_Work.tex
\section{Future Work}
\label{sec:future_work}
TS-Memory motivates a broader \emph{retrieval-to-memory} paradigm, where retrieval-induced distributional knowledge is compiled into a compact module for retrieval-free deployment. Future work can extend this paradigm along several directions.

\noindent\textbf{Continual and streaming memory updates.}
An appealing direction is enabling memory modules to evolve with newly observed data in non-stationary environments, while preserving retrieval-free inference.
This includes rolling-buffer teacher refresh, lightweight incremental distillation, and continual-learning regularizers that stabilize memory representations across drift.

\noindent\textbf{Richer teachers and reliability learning.}
Future work may explore stronger offline teacher signals, such as hybrid similarity metrics that combine embedding distance with correlation- or frequency-aware matching, as well as teacher ensembles across multiple retrieval spaces.
In addition, learned reliability predictors calibrated to downstream utility can provide a principled way to decide when teacher signals should guide memory learning.

\noindent\textbf{Broader alignment families for distribution shift.}
Beyond basic alignment settings, it is promising to study leakage-safe alignment families that capture a wider spectrum of non-stationarities, including scale changes, trend deformations, and temporal warping.
Such alignment can improve neighbor correspondence and yield sharper, more informative teacher distributions across domains.

\noindent\textbf{Input-adaptive fusion and uncertainty-aware calibration.}
Another direction is to make fusion between backbone and memory predictions context-conditioned, allowing the model to adapt memory reliance based on uncertainty and domain mismatch.
This could further improve probabilistic calibration by coupling fusion decisions with uncertainty-aware confidence modeling.

\noindent\textbf{Scalable teacher compilation and broader tasks.}
To support large-scale and multi-domain deployment, future work can develop more efficient teacher compilation pipelines, including approximate neighbor search, teacher-set distillation, and shared teacher banks across related domains.
Beyond long-horizon forecasting, retrieval-to-memory distillation may extend naturally to imputation, anomaly detection, and decision-focused forecasting, where retrieved analogs can provide privileged distributional supervision.

\input{tables/different-TSFMs-with-TS-Memory}

\input{tables/ragvsTS-Memory}

\input{tables/efficiency_query}

\input{tables/efficiency_time}

\input{tables/different-size-of-Chronos}

\input{tables/plugmem_merged_table}

\input{tables/TSMemoryLoRA-PlugMem-Size}

%% file: tables/different-TSFMs-with-TS-Memory.tex
\begin{table*}[!htpb]
\begin{center}
\caption{Full long-term forecasting results of TS-Memory across frozen TSFM backbones. Results are averaged over forecasting horizons $H \in $\{96, 192, 336, 720\}. Lower values indicate better performance. Best results are highlighted in \textbf{bold}, and second best results are underlined. $\Delta$ denotes the relative change (\%) of Avg after adding TS-Memory compared to Origin.}
\begin{small}
\setlength\tabcolsep{4pt}
\scalebox{0.95}{
\begin{tabular}{c|c|cc|cc|cc|cc|cc|cc|cc|cc}
\toprule
\multicolumn{2}{c|}{TSFM } &
\multicolumn{4}{c|}{ChronosBolt (base)} &
\multicolumn{4}{c|}{Chronos2 (base)} &
\multicolumn{4}{c|}{Sundial (base)} &
\multicolumn{4}{c}{TimesFM (base)} \\
\midrule
\multicolumn{2}{c|}{Dataset}  &
\multicolumn{2}{c|}{Origin} & \multicolumn{2}{c|}{TS-Memory} &
\multicolumn{2}{c|}{Origin} & \multicolumn{2}{c|}{TS-Memory} &
\multicolumn{2}{c|}{Origin} & \multicolumn{2}{c|}{TS-Memory} &
\multicolumn{2}{c|}{Origin} & \multicolumn{2}{c}{TS-Memory} \\
\midrule
\multicolumn{2}{c|}{Metric} &
MSE & MAE & MSE & MAE &
MSE & MAE & MSE & MAE &
MSE & MAE & MSE & MAE &
MSE & MAE & MSE & MAE \\
\midrule

\multirow{6}{*}{\rotatebox{90}{ETTh1}}
& 96  & \underline{0.3849} & \underline{0.3798} & \textbf{0.3736} & \textbf{0.3772} & \underline{0.3948} & \underline{0.3800} & \textbf{0.3704} & \textbf{0.3751} & \underline{0.3574} & \underline{0.3826} & \textbf{0.3540} & \textbf{0.3807} & \underline{0.4211} & \underline{0.4009} & \textbf{0.3920} & \textbf{0.3901} \\
& 192 & \underline{0.4417} & \underline{0.4122} & \textbf{0.4103} & \textbf{0.4071} & \underline{0.4382} & \underline{0.4073} & \textbf{0.4153} & \textbf{0.4019} & \underline{0.3920} & \underline{0.4056} & \textbf{0.3881} & \textbf{0.4026} & \underline{0.4721} & \underline{0.4323} & \textbf{0.4406} & \textbf{0.4209} \\
& 336 & \underline{0.4814} & \underline{0.4331} & \textbf{0.4393} & \textbf{0.4245} & \underline{0.4679} & \underline{0.4231} & \textbf{0.4394} & \textbf{0.4168} & \underline{0.4065} & \underline{0.4159} & \textbf{0.4038} & \textbf{0.4142} & \underline{0.5098} & \underline{0.4548} & \textbf{0.4578} & \textbf{0.4394} \\
& 720 & \underline{0.4836} & \underline{0.4489} & \textbf{0.4616} & \textbf{0.4451} & \underline{0.4672} & \underline{0.4355} & \textbf{0.4546} & \textbf{0.4331} & \underline{0.4424} & \underline{0.4453} & \textbf{0.4388} & \textbf{0.4430} & \underline{0.5140} & \underline{0.4813} & \textbf{0.5016} & \textbf{0.4752} \\
\bottomrule
\end{tabular}}
\end{small}

\label{tab:ts_plugmem_full}
\end{center}
\end{table*}

%% file: tables/ragvsTS-Memory.tex

\begin{table*}[!p]
\begin{center}
\caption{Full long-term forecasting results comparing TS-Memory with online retrieval baselines on ChronosBolt. We use the same protocol as in Table~\ref{tab:ts_plugmem_full}.}
\begin{small}
\setlength\tabcolsep{3.5pt}
\scalebox{1}{
\begin{tabular}{c|c|ccc|ccc|ccc|ccc|ccc}
\toprule
\multicolumn{2}{c|}{Dataset} &
\multicolumn{3}{c|}{Origin} &
\multicolumn{3}{c|}{RAFT} &
\multicolumn{3}{c|}{TS-RAG} &
\multicolumn{3}{c|}{LoRA} &
\multicolumn{3}{c}{TS-Memory} \\
\midrule
\multicolumn{2}{c|}{Metric} &
MSE & MAE & CRPS &
MSE & MAE & CRPS &
MSE & MAE & CRPS &
MSE & MAE & CRPS &
MSE & MAE & CRPS \\
\midrule
\multirow{6}{*}{\rotatebox{90}{ETTh1}}
& 96   & 0.3849 & 0.3798 & 0.3730 & 0.3782 & 0.3798 & \underline{0.3698} & 0.3780 & 0.3787 & 0.3708 & \underline{0.3764} & \underline{0.3783} & 0.3751 & \textbf{0.3736} & \textbf{0.3772} & \textbf{0.3634} \\
& 192  & 0.4417 & 0.4122 & 0.4177 & 0.4308 & 0.4106 & 0.4126 & 0.4319 & 0.4114 & 0.4161 & \underline{0.4254} & \underline{0.4102} & \underline{0.4111} & \textbf{0.4103} & \textbf{0.4071} & \textbf{0.3913} \\
& 336  & 0.4814 & 0.4331 & 0.4590 & 0.4648 & 0.4299 & 0.4520 & 0.4699 & 0.4337 & 0.4599 & \underline{0.4549} & \underline{0.4292} & \underline{0.4402} & \textbf{0.4393} & \textbf{0.4245} & \textbf{0.4133} \\
& 720  & 0.4836 & 0.4489 & 0.4980 & 0.4734 & 0.4501 & 0.4942 & 0.4893 & 0.4619 & 0.5148 & \textbf{0.4568} & \underline{0.4472} & \underline{0.4869} & \underline{0.4616} & \textbf{0.4451} & \textbf{0.4606} \\
\cmidrule{2-17}
& Avg  & 0.4479 & 0.4185 & 0.4369 & 0.4368 & 0.4176 & 0.4321 & 0.4423 & 0.4214 & 0.4404 & \underline{0.4284} & \underline{0.4162} & \underline{0.4283} & \textbf{0.4212} & \textbf{0.4135} & \textbf{0.4071} \\
& $\Delta$ & -- & -- & -- & -2.48\% & -0.22\% & -1.10\% & -1.25\% & +0.69\% & +0.80\% & -4.35\% & -0.55\% & -1.97\% & -5.96\% & -1.19\% & -6.82\% \\
\bottomrule
\end{tabular}}
\end{small}

\label{tab:chronosbolt_base_ablation}
\end{center}
\end{table*}

%% file: tables/efficiency_query.tex
\begin{table*}[!htbp]
\begin{center}
\caption{Full inference latency breakdown (ms/query) for Origin, online retrieval baselines, LoRA, and TS-Memory on ChronosBolt. We report retrieval time (Retr), forward time (Fwd), total latency (Total), and retrieval fraction (Frac). $\Delta$ denotes the relative change (\%) of Avg compared to Origin. We use the same protocol as in Table~\ref{tab:ts_plugmem_full}.}
\begin{small}
\setlength\tabcolsep{3pt}
\scalebox{0.92}{
\begin{tabular}{c|c|cccc|cccc|cccc|cccc|cccc}
\toprule

\multicolumn{2}{c|}{Dataset} &
\multicolumn{4}{c|}{Origin} &
\multicolumn{4}{c|}{RAFT} &
\multicolumn{4}{c|}{TS-RAG} &
\multicolumn{4}{c|}{LoRA} &
\multicolumn{4}{c}{TS-Memory} \\
\midrule
\multicolumn{2}{c|}{Metric} &
Retr & Fwd & Total & Frac &
Retr & Fwd & Total & Frac &
Retr & Fwd & Total & Frac &
Retr & Fwd & Total & Frac &
Retr & Fwd & Total & Frac \\
\midrule
\multirow{6}{*}{\rotatebox{90}{ETTh1}} & 96 & 0 & 0.914 & 0.914 & 0\% & 0.292 & \underline{0.997} & 1.289 & 22.7\% & 1.262 & 1.025 & 2.287 & 55.2\% & \underline{0} & 1.217 & \underline{1.217} & \underline{0\%} & \textbf{0} & \textbf{0.980} & \textbf{0.980} & \textbf{0\%} \\
 & 192 & 0 & 1.433 & 1.433 & 0\% & 0.383 & \underline{1.587} & 1.970 & 19.4\% & 1.835 & 1.590 & 3.425 & 53.6\% & \underline{0} & 1.727 & \underline{1.727} & \underline{0\%} & \textbf{0} & \textbf{1.544} & \textbf{1.544} & \textbf{0\%} \\
 & 336 & 0 & 3.297 & 3.297 & 0\% & 1.062 & 3.550 & 4.612 & 23.0\% & 3.597 & \underline{3.537} & 7.134 & 50.4\% & \underline{0} & 4.051 & \underline{4.051} & \underline{0\%} & \textbf{0} & \textbf{3.463} & \textbf{3.463} & \textbf{0\%} \\
 & 720 & 0 & 8.464 & 8.464 & 0\% & 1.618 & 9.221 & 10.839 & 14.9\% & 7.436 & \underline{9.054} & 16.489 & 45.1\% & \underline{0} & 10.133 & \underline{10.133} & \underline{0\%} & \textbf{0} & \textbf{8.776} & \textbf{8.776} & \textbf{0\%} \\
\cmidrule{2-22}
 & Avg & 0 & 3.527 & 3.527 & 0\% & 0.839 & 3.839 & 4.677 & 20.0\% & 3.533 & \underline{3.801} & 7.334 & 51.1\% & \underline{0} & 4.282 & \underline{4.282} & \underline{0\%} & \textbf{0} & \textbf{3.691} & \textbf{3.691} & \textbf{0\%} \\
 & $\Delta$ & -- & -- & -- & -- & -- & +8.84\% & +32.62\% & -- & -- & +7.77\% & +107.94\% & -- & -- & +21.41\% & +21.41\% & -- & -- & +4.65\% & +4.65\% & -- \\
\bottomrule
\end{tabular}}
\end{small}
\label{tab:chronosbolt_time_latency}
\end{center}
\end{table*}

%% file: tables/efficiency_time.tex
\begin{table*}[!p]
\begin{center}
\caption{Full end-to-end inference time breakdown (s) for Origin, online retrieval baselines, LoRA, and TS-Memory on ChronosBolt. We report retrieval time (Retr), forward time (Fwd), total latency (Total), and retrieval fraction (Frac). $\Delta$ denotes the relative change (\%) of Avg compared to Origin. We use the same protocol as in Table~\ref{tab:ts_plugmem_full}.}
\begin{small}
\setlength\tabcolsep{3pt}
\scalebox{0.88}{
\begin{tabular}{c|c|cccc|cccc|cccc|cccc|cccc}
\toprule

\multicolumn{2}{c|}{Dataset} &
\multicolumn{4}{c|}{Origin} &
\multicolumn{4}{c|}{RAFT} &
\multicolumn{4}{c|}{TS-RAG} &
\multicolumn{4}{c|}{LoRA} &
\multicolumn{4}{c}{TS-Memory} \\
\midrule
\multicolumn{2}{c|}{Metric} &
Retr & Fwd & Total & Frac &
Retr & Fwd & Total & Frac &
Retr & Fwd & Total & Frac &
Retr & Fwd & Total & Frac &
Retr & Fwd & Total & Frac \\
\midrule
\multirow{6}{*}{\rotatebox{90}{ETTh1}} & 96 & 0 & 14.5 & 14.5 & 0\% & 4.2 & \textbf{14.3} & 18.5 & 22.7\% & 18.1 & \underline{14.7} & 32.9 & 55.2\% & \underline{0} & 18.3 & \underline{18.3} & \underline{0\%} & \textbf{0} & 15.6 & \textbf{15.6} & \textbf{0\%} \\
 & 192 & 0 & 21.8 & 21.8 & 0\% & 5.2 & \textbf{21.7} & 27.0 & 19.4\% & 25.1 & \underline{21.8} & 46.9 & 53.6\% & \underline{0} & 24.8 & \underline{24.8} & \underline{0\%} & \textbf{0} & 23.5 & \textbf{23.5} & \textbf{0\%} \\
 & 336 & 0 & 46.9 & 46.9 & 0\% & 13.5 & \textbf{45.1} & 58.5 & 23.0\% & 45.7 & \underline{44.9} & 90.6 & 50.4\% & \underline{0} & 54.0 & \underline{54.0} & \underline{0\%} & \textbf{0} & 49.3 & \textbf{49.3} & \textbf{0\%} \\
 & 720 & 0 & 97.7 & 97.7 & 0\% & 16.2 & \underline{92.3} & 108.5 & 14.9\% & 74.4 & \textbf{90.6} & 165.0 & 45.1\% & \underline{0} & 107.9 & \underline{107.9} & \underline{0\%} & \textbf{0} & 101.3 & \textbf{101.3} & \textbf{0\%} \\
\cmidrule{2-22}
 & Avg & 0 & 45.2 & 45.2 & 0\% & 9.8 & \underline{43.4} & 53.1 & 20.0\% & 40.8 & \textbf{43.0} & 83.8 & 51.1\% & \underline{0} & 51.2 & \underline{51.2} & \underline{0\%} & \textbf{0} & 47.4 & \textbf{47.4} & \textbf{0\%} \\
 & $\Delta$ & -- & -- & -- & -- & -- & -4.2\% & +17.4\% & -- & -- & -5.0\% & +85.3\% & -- & -- & +13.4\% & +13.4\% & -- & -- & +4.8\% & +4.8\% & -- \\
\bottomrule
\end{tabular}}
\end{small}
\label{tab:chronosbolt_total_time}
\end{center}
\end{table*}

%% file: tables/different-size-of-Chronos.tex
\begin{table*}[!htbp]
\begin{center}
\caption{Full scaling study of TS-Memory across ChronosBolt model sizes. We use the same protocol as in Table~\ref{tab:ts_plugmem_full}.}
\begin{small}
\setlength\tabcolsep{4pt}
\scalebox{0.95}{
\begin{tabular}{c|c|cc|cc|cc|cc|cc|cc|cc|cc}
\toprule
\multicolumn{2}{c|}{Type} &
\multicolumn{4}{c|}{ChronosBolt Base (205M)} &
\multicolumn{4}{c|}{ChronosBolt Small (48M)} &
\multicolumn{4}{c|}{ChronosBolt Mini (21M)} &
\multicolumn{4}{c}{ChronosBolt Tiny (9M)} \\
\midrule
\multicolumn{2}{c|}{Dataset} &
\multicolumn{2}{c|}{Origin} & \multicolumn{2}{c|}{TS-Memory} &
\multicolumn{2}{c|}{Origin} & \multicolumn{2}{c|}{TS-Memory} &
\multicolumn{2}{c|}{Origin} & \multicolumn{2}{c|}{TS-Memory} &
\multicolumn{2}{c|}{Origin} & \multicolumn{2}{c}{TS-Memory} \\
\midrule
\multicolumn{2}{c|}{Metric} &
MSE & MAE & MSE & MAE &
MSE & MAE & MSE & MAE &
MSE & MAE & MSE & MAE &
MSE & MAE & MSE & MAE \\
\midrule

\multirow{6}{*}{\rotatebox{90}{ETTh1}}
& 96 & \underline{0.3849} & \underline{0.3798} & \textbf{0.3736} & \textbf{0.3772} & \underline{0.3941} & \underline{0.3816} & \textbf{0.3802} & \textbf{0.3782} & \underline{0.3801} & \underline{0.3788} & \textbf{0.3717} & \textbf{0.3769} & \underline{0.3844} & \underline{0.3821} & \textbf{0.3731} & \textbf{0.3796} \\
& 192 & \underline{0.4417} & \underline{0.4122} & \textbf{0.4103} & \textbf{0.4071} & \underline{0.4551} & \underline{0.4141} & \textbf{0.4248} & \textbf{0.4081} & \underline{0.4354} & \underline{0.4103} & \textbf{0.4124} & \textbf{0.4065} & \underline{0.4381} & \underline{0.4121} & \textbf{0.4137} & \textbf{0.4079} \\
& 336 & \underline{0.4814} & \underline{0.4331} & \textbf{0.4393} & \textbf{0.4245} & \underline{0.4996} & \underline{0.4362} & \textbf{0.4484} & \textbf{0.4267} & \underline{0.4770} & \underline{0.4336} & \textbf{0.4365} & \textbf{0.4256} & \underline{0.4785} & \underline{0.4340} & \textbf{0.4377} & \textbf{0.4262} \\
\bottomrule
\end{tabular}}
\end{small}

\label{tab:chronosbolt_ts_plugmem_full}
\end{center}
\end{table*}

%% file: tables/plugmem_merged_table.tex
\begin{table*}[!p]
\begin{center}
\caption{Full cross-model transfer results of TS-Memory across retrieval teachers and frozen TSFM backbones across horizons. We use the same protocol as in Table~\ref{tab:ts_plugmem_full}.}
\begin{small}
\setlength\tabcolsep{2pt}
\scalebox{0.74}{%
\begin{tabular}{c|c|cc|cc|cc|cc|cc|cc||cc|cc|cc|cc|cc|cc}
\toprule
\multicolumn{2}{c|}{Type} & \multicolumn{12}{c||}{Chronos2-PlugMem} & \multicolumn{12}{c}{Sundial-PlugMem} \\
\midrule
\multicolumn{2}{c|}{Dataset} &
\multicolumn{4}{c|}{ChronosBolt (base)} &
\multicolumn{4}{c|}{Sundial (base)} &
\multicolumn{4}{c||}{TimesFM (base)} &
\multicolumn{4}{c|}{ChronosBolt (base)} &
\multicolumn{4}{c|}{Chronos2 (base)} &
\multicolumn{4}{c}{TimesFM (base)} \\
\midrule
\multicolumn{2}{c|}{} &
\multicolumn{2}{c|}{Origin} & \multicolumn{2}{c|}{TS-Memory} &
\multicolumn{2}{c|}{Origin} & \multicolumn{2}{c|}{TS-Memory} &
\multicolumn{2}{c|}{Origin} & \multicolumn{2}{c||}{TS-Memory} &
\multicolumn{2}{c|}{Origin} & \multicolumn{2}{c|}{TS-Memory} &
\multicolumn{2}{c|}{Origin} & \multicolumn{2}{c|}{TS-Memory} &
\multicolumn{2}{c|}{Origin} & \multicolumn{2}{c}{TS-Memory} \\
\midrule
\multicolumn{2}{c|}{Metric} &
MSE & MAE & MSE & MAE &
MSE & MAE & MSE & MAE &
MSE & MAE & MSE & MAE &
MSE & MAE & MSE & MAE &
MSE & MAE & MSE & MAE &
MSE & MAE & MSE & MAE \\
\midrule

\multirow{6}{*}{\rotatebox{90}{ETTh1}} & 96 & \underline{0.3849} & \underline{0.3798} & \textbf{0.3586} & \textbf{0.3794} & \underline{0.3574} & \underline{0.3826} & \textbf{0.3489} & \textbf{0.3791} & \underline{0.4211} & \underline{0.4009} & \textbf{0.3694} & \textbf{0.3900} & \underline{0.3849} & \textbf{0.3798} & \textbf{0.3587} & \underline{0.3812} & \underline{0.3948} & \textbf{0.3800} & \textbf{0.3646} & \underline{0.3804} & \underline{0.4211} & \underline{0.4009} & \textbf{0.3747} & \textbf{0.3892} \\
\bottomrule
\end{tabular}}
\end{small}
\label{tab:plugmem_merged}
\end{center}
\end{table*}

%% file: tables/TSMemoryLoRA-PlugMem-Size.tex
\begin{table*}[!t]
\centering

\caption{Full long-term forecasting results under different train-test domain configurations (Left: TS-Memory transfer; Right: LoRA). We use the same protocol as in Table~\ref{tab:ts_plugmem_full}.}
\label{tab:domain_split_lora_tsmemory}

{\scriptsize
\setlength{\tabcolsep}{2pt}
\renewcommand{\arraystretch}{1}

\scalebox{1.2}{%
\begin{tabular}{c|c|cc|cc|cc|cc||cc|cc|cc|cc}
\toprule
\multicolumn{2}{c|}{} & \multicolumn{8}{c||}{\textbf{TS-Memory}} & \multicolumn{8}{c}{\textbf{LoRA}} \\
\midrule
\multicolumn{2}{c|}{Dataset} &
\multicolumn{2}{c|}{ETTh1} &
\multicolumn{2}{c|}{ETTh2} &
\multicolumn{2}{c|}{ETTm1} &
\multicolumn{2}{c||}{ETTm2} &
\multicolumn{2}{c|}{ETTh1} &
\multicolumn{2}{c|}{ETTh2} &
\multicolumn{2}{c|}{ETTm1} &
\multicolumn{2}{c}{ETTm2} \\
\midrule
\multicolumn{2}{c|}{Metric} &
MSE & MAE & MSE & MAE & MSE & MAE & MSE & MAE &
MSE & MAE & MSE & MAE & MSE & MAE & MSE & MAE \\
\midrule

\multirow{5}{*}{\rotatebox{90}{Origin}}
& 96  & 0.3849 & 0.3798 & 0.2895 & 0.3251 & 0.3319 & 0.3332 & 0.1777 & 0.2448
     & 0.3849 & 0.3798 & 0.2895 & 0.3251 & 0.3319 & 0.3332 & 0.1777 & 0.2448 \\
& 192 & 0.4417 & 0.4122 & 0.3598 & 0.3695 & 0.3870 & 0.3658 & 0.2511 & 0.2935
     & 0.4417 & 0.4122 & 0.3598 & 0.3695 & 0.3870 & 0.3658 & 0.2511 & 0.2935 \\
& 336 & 0.4814 & 0.4331 & 0.4029 & 0.4025 & 0.4363 & 0.3934 & 0.3170 & 0.3359
     & 0.4814 & 0.4331 & 0.4029 & 0.4025 & 0.4363 & 0.3934 & 0.3170 & 0.3359 \\
& 720 & 0.4836 & 0.4489 & 0.4140 & 0.4210 & 0.5279 & 0.4391 & 0.4160 & 0.3950
     & 0.4836 & 0.4489 & 0.4140 & 0.4210 & 0.5279 & 0.4391 & 0.4160 & 0.3950 \\
\cmidrule{2-10}\cmidrule{11-18}
& Avg & 0.4479 & 0.4185 & 0.3666 & 0.3795 & 0.4208 & 0.3829 & 0.2905 & 0.3173
     & 0.4479 & 0.4185 & 0.3665 & 0.3795 & 0.4208 & 0.3829 & 0.2904 & 0.3173 \\
\bottomrule
\end{tabular}}%
} 

\vspace{2em}

\caption{Full scaling study of PlugMem sizes on ChronosBolt. We use the same protocol as in Table~\ref{tab:ts_plugmem_full}.}
\label{tab:mem-size}

{\small
\setlength\tabcolsep{2.2pt}
\renewcommand{\arraystretch}{0.95}
\scalebox{0.9}{%
\begin{tabular}{l|c|cc|cc|cc|cc|cc|cc|cc|cc}
\toprule
\multirow{2}{*}{Method} & \multirow{2}{*}{H} &
\multicolumn{2}{c|}{ETTh1} &
\multicolumn{2}{c|}{ETTh2} &
\multicolumn{2}{c|}{ETTm1} &
\multicolumn{2}{c|}{ETTm2} &
\multicolumn{2}{c|}{Electricity} &
\multicolumn{2}{c|}{Exchange-rate} &
\multicolumn{2}{c|}{Traffic} &
\multicolumn{2}{c}{Weather} \\
\cmidrule(lr){3-4}\cmidrule(lr){5-6}\cmidrule(lr){7-8}\cmidrule(lr){9-10}
\cmidrule(lr){11-12}\cmidrule(lr){13-14}\cmidrule(lr){15-16}\cmidrule(lr){17-18}
& & MSE & MAE & MSE & MAE & MSE & MAE & MSE & MAE & MSE & MAE & MSE & MAE & MSE & MAE & MSE & MAE \\
\midrule

\multirow{5}{*}{Origin}
& 96  & 0.3849 & 0.3798 & 0.2895 & 0.3251 & 0.3319 & 0.3332 & 0.1777 & 0.2448 & 0.1242 & 0.2105 & 0.0992 & 0.2191 & 0.3870 & 0.2409 & 0.1775 & 0.2102 \\
& 192 & 0.4417 & 0.4122 & 0.3598 & 0.3695 & 0.3870 & 0.3658 & 0.2511 & 0.2935 & 0.1428 & 0.2287 & 0.1944 & 0.3140 & 0.4137 & 0.2531 & 0.2242 & 0.2535 \\
& 336 & 0.4814 & 0.4331 & 0.4029 & 0.4025 & 0.4363 & 0.3934 & 0.3170 & 0.3359 & 0.1613 & 0.2482 & 0.3420 & 0.4238 & 0.4355 & 0.2633 & 0.2838 & 0.2947 \\
& 720 & 0.4836 & 0.4489 & 0.4140 & 0.4210 & 0.5279 & 0.4391 & 0.4160 & 0.3950 & 0.2069 & 0.2882 & 0.8493 & 0.6915 & 0.5033 & 0.2934 & 0.3670 & 0.3465 \\
\cmidrule{2-18}
& AVG & 0.4479 & 0.4185 & 0.3665 & 0.3795 & 0.4208 & 0.3829 & 0.2904 & 0.3173 & 0.1588 & 0.2439 & 0.3712 & 0.4121 & 0.4349 & 0.2627 & 0.2631 & 0.2762 \\
\midrule

\multirow{5}{*}{\makecell{PlugMem\\Small (2.5M)}}
& 96  & \textbf{0.3684} & \textbf{0.3770} & 0.2854 & \underline{0.3238} & \textbf{0.2978} & \textbf{0.3226} & \textbf{0.1643} & \textbf{0.2387} & \underline{0.1207} & \underline{0.2083} & 0.0972 & 0.2168 & \underline{0.3813} & 0.2387 & \textbf{0.1536} & \textbf{0.1969} \\
\bottomrule
\end{tabular}
}}%

\end{table*}